\documentclass[12pt]{article}
\usepackage{latexsym,epsfig,graphicx,epstopdf,amsmath,amssymb,amscd,undertilde,multirow,chicago,psfrag,paralist,dsfont,url}
\usepackage[titletoc]{appendix}
\newcommand{\thetavec}{{\boldsymbol{\theta}}}

\newcommand{\Indfun}{{\mathds{1}}}

\newcommand{\thetavechat}{\widehat{\thetavec}}

\newcommand{\wh}{\widehat}

\newcommand{\xvec}{\boldsymbol{x}}

\begin{document}

\title{Reliability Analysis of Artificial Intelligence Systems Using Recurrent Events Data from Autonomous Vehicles}

\author{
Yili Hong$^1$, Jie Min$^1$, Caleb B. King$^2$, and William Q. Meeker$^3$\\[1.5ex]
{\small $^1$Department of Statistics, Virginia Tech, Blacksburg, VA 24061}\\
{\small $^2$JMP Division, SAS, Cary, NC 27513}\\
{\small $^3$Department of Statistics, Iowa State University, Ames, IA 50011}
}

\date{}

\maketitle
\begin{abstract}
Artificial intelligence (AI) systems have become increasingly common and the trend will continue. Examples of AI systems include autonomous vehicles (AV), computer vision, natural language processing, and AI medical experts. To allow for safe and effective deployment of AI systems, the reliability of such systems needs to be assessed. Traditionally, reliability assessment is based on reliability test data and the subsequent statistical modeling and analysis. The availability of reliability data for AI systems, however, is limited because such data are typically sensitive and proprietary. The California Department of Motor Vehicles (DMV) oversees and regulates an AV testing program, in which many AV manufacturers are conducting AV road tests. Manufacturers participating in the program are required to report recurrent disengagement events to California DMV. This information is being made available to the public. In this paper, we use recurrent disengagement events as a representation of the reliability of the AI system in AV, and propose a statistical framework for modeling and analyzing the recurrent events data from AV driving tests. We use traditional parametric models in software reliability and propose a new nonparametric model based on monotonic splines to describe the event process. We develop inference procedures for selecting the best models, quantifying uncertainty, and testing heterogeneity in the event process. We then analyze the recurrent events data from four AV manufacturers, and make inferences on the reliability of the AI systems in AV. We also describe how the proposed analysis can be applied to assess the reliability of other AI systems.

\textbf{Key Words:} Disengagement Events; Fractional Random Weight Bootstrap; Gompertz Model; Monotonic Splines;  Software Reliability; Self-driving Cars.
\end{abstract}

\newpage

\section{Introduction}
\subsection{The Problem}
With the rapid development of new technology, artificial intelligence (AI) systems are emerging in many areas. Typical applications of AI systems include autonomous vehicles (AV), computer vision, speech recognition, and AI medical experts. The reliability and safety of AI systems need to be assessed before massive deployment in the field. For example, the reliability of AV needs to be demonstrated so that people can use them with confidence. Traditionally, reliability assessment of products and systems is based on reliability test data collected from the laboratory and the field. Reliability information is then extracted from statistical modeling and analysis of the data.

Commonly used data types for reliability analysis are time-to-event data, degradation data, and recurrent events data. Reliability data collected by manufacturers are highly sensitive and are usually not publicly available. In the area of AV, however,  the \citeANP{CAdriving} (DMV) has launched an AV driving program. More details of the study are given in Section~\ref{sec:CA.driving.study}. Many AV manufacturers are participating in the program and are conducting AV road test in California. Manufacturers are required to report disengagement events and mileage information to the California DMV. The reported events are available for public access. Because of the availability of these recurrent events data, we focus on the reliability analysis of the AI systems in AV units in this paper.

A disengagement event happens when the AI system and/or the backup driver determines that the driver needs to take over the driving. The recurrence rates of disengagement events can be used as a proxy for the reliability of the AI system in AV. A lower occurrence rate (event rate) of disengagement events would indicate a more reliable AI system in the AV. In the reliability literature, parametric forms have been used to describe the event rate through a nonhomogeneous Poisson process (NHPP) model. In practice, some specific questions arise in the analysis of the recurrent disengagement events data,

\begin{inparaitem}
\item How to model the event process, and what kind of parametric forms should be used?

\item Does the parametric form provide an adequate fit to the data, and are there any other flexible forms for modeling?

\item Is there any population heterogeneity in the event processes from multiple test units?
\end{inparaitem}
To answer those practical questions, we develop a statistical framework for modeling and analyzing the recurrent events data from AV driving tests.

Specifically, we use the NHPP model to describe the disengagement event process with adjustment for the time-varying mileage data using traditional parametric models in software reliability for the intensity functions. We also propose a new nonparametric model based on monotonic splines to describe the event process. The spline model is flexible enough to fit various patterns in the event process and can also be used to assess if the fully-parametric model provides an adequate fit. We develop inference procedures for selecting the best models and quantifying uncertainty using the fractional-random-weight bootstrap. The parametric models and spline models can be used as complementary tools. In addition, we use the gamma frailty model to quantify and assess heterogeneity in the event process.

From the California driving study, we use data from four manufacturers that have been conducting extensive AV driving tests in California. We apply the developed methods to analyze the recurrent events data from the four AV manufacturers. Based on the modeling, we make inferences on the reliability of the AI systems in AV, and summarize interesting findings on the reliability of the AI systems in AV. Although our analysis focuses on the AI systems in AV, the statistical analysis can also be applied to assess the reliability of other AI systems.

\subsection{Literature Review}
There is only a small amount of literature on reliability analysis of AI systems. \shortciteN{Amodeietal2016} provided a general discussion on AI safety and outlined five concrete areas for AI safety research.  \shortciteN{Bosioetal2019} conducted a reliability analysis of deep convolutional neural network (CNN) developed for automotive applications using fault injections. \shortciteN{Goldsteinetal2020} investigated the impact of transient faults on the reliability of compressed deep CNN. \shortciteN{zhao2020safety} proposed a safety framework based on Bayesian inference for critical systems using deep learning models. \citeN{AlshemaliKalita2020} provided a review on improving the reliability of natural language processing. Due to the limited availability of test data, statistical reliability analysis of AI reliability/safety is in an emerging stage.

As an example of the modeling of the reliability and safety of AV, \citeN{kalra2016driving} used a statistical hypothesis testing approach to determine the needed miles of driving to demonstrate AV safety. \citeN{Asljungetal2017} used extreme value theory to model the safety of AV. \shortciteN{michelmore2019uncertainty} designed a statistical framework to evaluate the safety of deep neural network controllers and assessed the safety of AV. \shortciteN{Burtonetal2020} provided a multi-disciplinary perspective on the safety of AV from engineering, ethics, and law aspects. Most existing modeling framework, however, does not involve large scale field-testing data, which is essential for reliability assessment.

The California driving test data provide unique opportunities for data analysis. Regarding the analysis of the California driving data, \citeN{dixit2016autonomous} and \citeN{favaro2018autonomous} analyzed the causes of disengagement events using the California driving test data up to 2017 and showed the relationship between disengagement events per miles and cumulative miles. \shortciteN{Lvetal2018} performed a descriptive analysis of the causes of disengagement events using the California driving test data from 2014 to 2015, and concluded that software issues and limitations were the most common reasons for disengagement events. \shortciteN{banerjee2018hands} used linear regressions to model the relationship between disengagement per mile and cumulative miles using the data from 2016 to 2017. \citeN{Merkel2018} conducted data analysis on the California driving data from 2015 to 2017 with aggregated counts data and least-squares fit. \shortciteN{Zhaoetal2019} proposed a general conservative Bayesian inference method to estimate the rate of events (crashes and fatality) and illustrated it with the California driving test data. \citeN{Boggsetal2020} did an exploratory analysis for AV crashes from the California driving study. So far, there is no comprehensive statistical treatment for the analysis of AI reliability and especially for AV reliability. Starting in 2018, the exact disengagement event time can be extracted from the California DMV report, which makes it possible to apply recurrent events modeling techniques.

In the reliability literature, NHPP models are widely used to analyze recurrent events. \citeN{ZuoMeekerWu2008}, and \citeN{HongLiOsborn2015} analyzed recurrent events data with window-observations, which has similar data types with the disengagement events data from the California driving study. Parametric models, such as the Musa-Okumoto model (e.g., \citeNP{musa1984logarithmic}) and the Gompertz model (e.g., \citeNP{huang2003unified}), are commonly used in software reliability applications (e.g., \citeNP{Wood1996}). \shortciteN{Ehrlichetal1998} used accelerated testing methods to study software reliability. \citeN{Burkeetal2020} proposed flexible parametric models for time-to-event data analysis. Useful reference books for reliability data analysis for researchers in the AI reliability area include \citeN{meekerescobar1998}, and \citeN{lawless2003}. Overall, parametric models are common in analyzing recurrent events data in the context of reliability study.

We propose to use monotonic splines (\citeNP{Ramsay1988}, and \citeNP{Meyer2008}) as a nonparametric method to model the event process. Although monotonic splines are used in some degradation settings (\shortciteNP{Xieetal2018}), the application of monotonic splines to recurrent events in reliability is new. To model population heterogeneity, the gamma frailty model is used. An early use of the gamma frailty model in reliability is found in \citeN{Lawless1995}. More recently, \citeN{ShanHongMeeker2020} used the gamma frailty model to describe seasonal warrant return data.  \citeN{DuchateauJanssen2008} provide a comprehensive review of frailty models.

Due to the complicated structure of the window-observed recurrent events data in our study, we use fractional-random-weight bootstrap as a convenient way to generate bootstrap samples for statistical inference. The idea of fractional-random-weight bootstrap is introduced in \citeN{rubin1981bayesian}, and some theoretical properties are shown in \citeN{JinYingWei2001}. \shortciteN{Xuetal2020} demonstrated the use of fractional-random-weight bootstrap in many complex applications in reliability, survival analysis, and logistic regression. A simultaneous confidence band (SCB) can be used to assess if a parametric model is adequate for fitting the data. \citeN{HongEscobarMeeker2010} showed that an SCB could be obtained from a simultaneous confidence region (SCR) for parameters. However, in the context of bootstrap, it is not straightforward to construct SCR for parameter estimators with multiple dimensions. Hence, we use the idea of the equal-precision band in \citeN{Nair1984} and use bootstrap samples to calibrate pointwise confidence intervals to provide SCB to quantify statistical uncertainty.

In summary, we provide a general analytic framework by integrating existing methods and proposing new methods for reliability analysis of data from an AI study. The parametric and nonparametric models, and the statistical interval and testing procedures will be useful tools for practitioners working in the area of AI reliability.

\subsection{Overview}
The rest of the paper is organized as follows. Section~\ref{sec:CA.driving.study} describes the California autonomous vehicle driving study and introduces the data. Section~\ref{sec:stat.model} describes the parametric model, the spline model, and the gamma frailty model that are used to describe the recurrent disengagement events data.  Section~\ref{sec:model.est.inf} describes the parameter estimation procedures and the inference procedures for various models. Section~\ref{sec:simulation.study} conducts a simulation study to show the statistical performance of the estimation procedures. Section~\ref{sec:data.analysis} conducts the data analysis and summarizes interesting findings. Section~\ref{sec:conclusoin} contains some concluding remarks and areas for future research.

\section{The California Autonomous Vehicles Driving Study}\label{sec:CA.driving.study}
\subsection{The Study}
This paper presents reliability modeling and analysis of autonomous vehicles (AV) using data from the California Department of Motor Vehicles (DMV) autonomous vehicle tester program, which has been in operation since 2014. The tester program allows manufacturers to test AV on California public roads with a human in the driver seat who can take control of the vehicle if necessary. Up to July 1, 2020, 62 manufacturers are permitted to perform AV drive testing. Manufacturers are required to report disengagement events annually, and collision of AV within 10 days of the accident. Before December 1, 2017, only the aggregated number of disengagement events per month was reported, while after that date, the exact date of the event is now reported. Thus, we focus on the analysis of the data after December 1, 2017.

Since it can be difficult to determine responsibility in collisions, a disengagement event is typically used as an alternative to determining unsafe auto-driving in the literature (e.g., \citeNP{Merkel2018}). During the period from December 1, 2017, to November 30, 2019, 34 manufacturers reported disengagement events. We use the data from disengagement events reported by \citeANP{Waymo}, \citeANP{Cruise}, \citeANP{PonyAI}, and \citeANP{Zoox} as these four manufacturers performed extensive on-road testing during this time period.

Here we provide a brief introduction to those four manufacturers. Waymo began as the Google Self-Driving Car Project in 2009, testing autonomous vehicles on public roads across six states in the United States.  Waymo joined the California tester program in 2015. Cruise is an autonomous vehicle company founded in 2013. It joined the California tester program in 2016 and tested AV in the urban environment of San Francisco. Pony AI was founded in 2016, developing autonomous driving technology globally. Pony AI started AV testing on California public roads in June 2017. Zoox is an autonomous vehicle company founded in 2013. They joined the California tester program in 2017 and tested AV in downtown San Francisco.

\subsection{Disengagement Events Data}
The California DMV requires manufacturers to report when their vehicles disengaged from autonomous mode during tests. A disengagement event occurs when there is an autonomous technology failure, or when situations requiring the test driver to take manual control of the vehicle to operate safely. Disengagement events can be initialized by a warning from the autonomous vehicle system, or by test drivers as the driver thinks it is not safe to continue auto driving. Disengagement reports are provided annually, and include the ID number of vehicles in testing, location and date of disengagement events, description of causing of disengagement, monthly autonomous mileage of each testing vehicle, and annual total autonomous miles of each testing vehicle.

The starting date for the data used for this paper is December 1, 2017. The study period is from December 1, 2017, to November 30, 2019, which is a 24 month or 2 year study period. The data for the period from December 1, 2018, to November 30, 2019, are available in csv format from the California DMV website, while the data for the period from December 1, 2017, to November 30, 2018, are available in pdf format that need to be manually converted to csv format.

\begin{figure}
\centering
\begin{tabular}{cc}
\includegraphics[width=.48\textwidth]{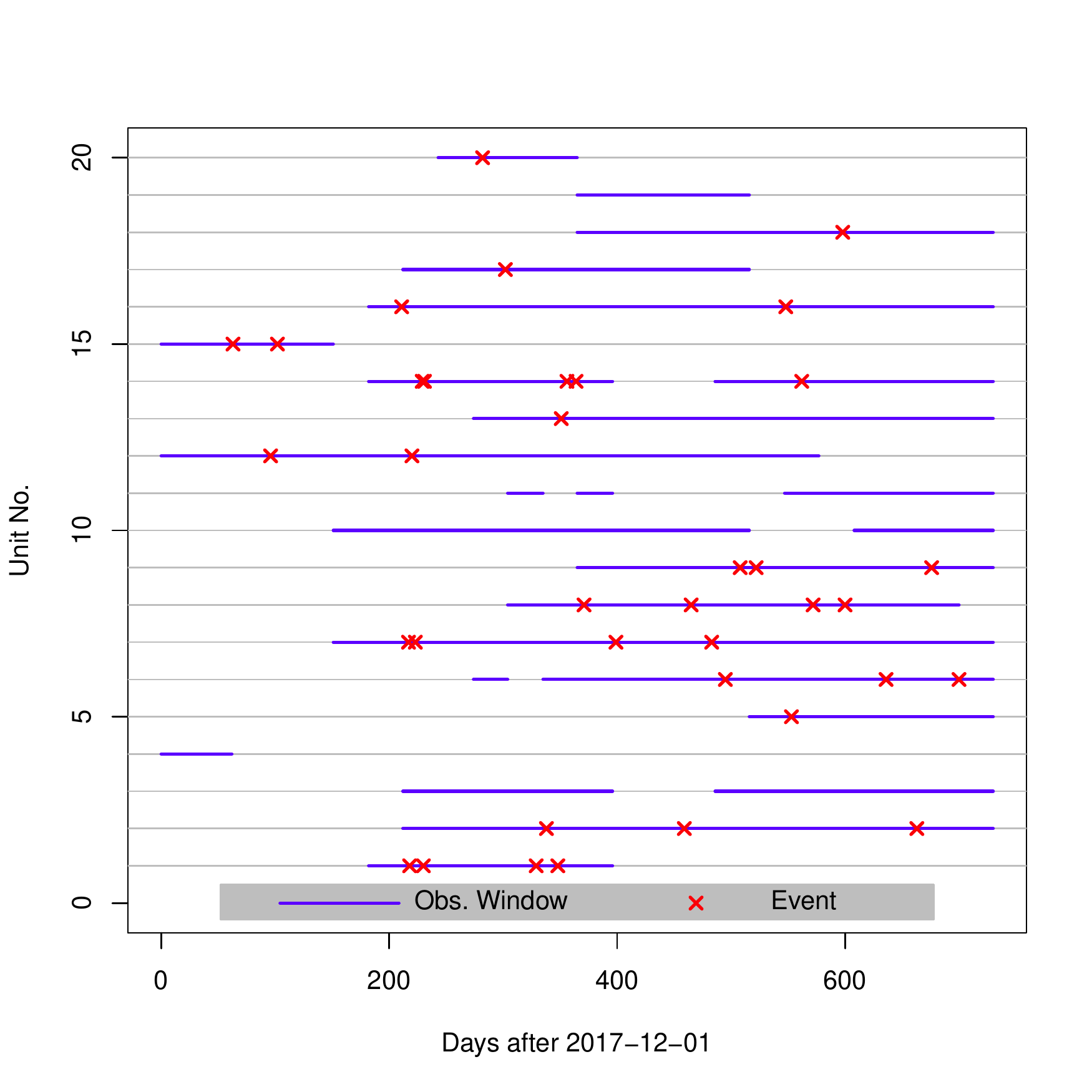}&
\includegraphics[width=.48\textwidth]{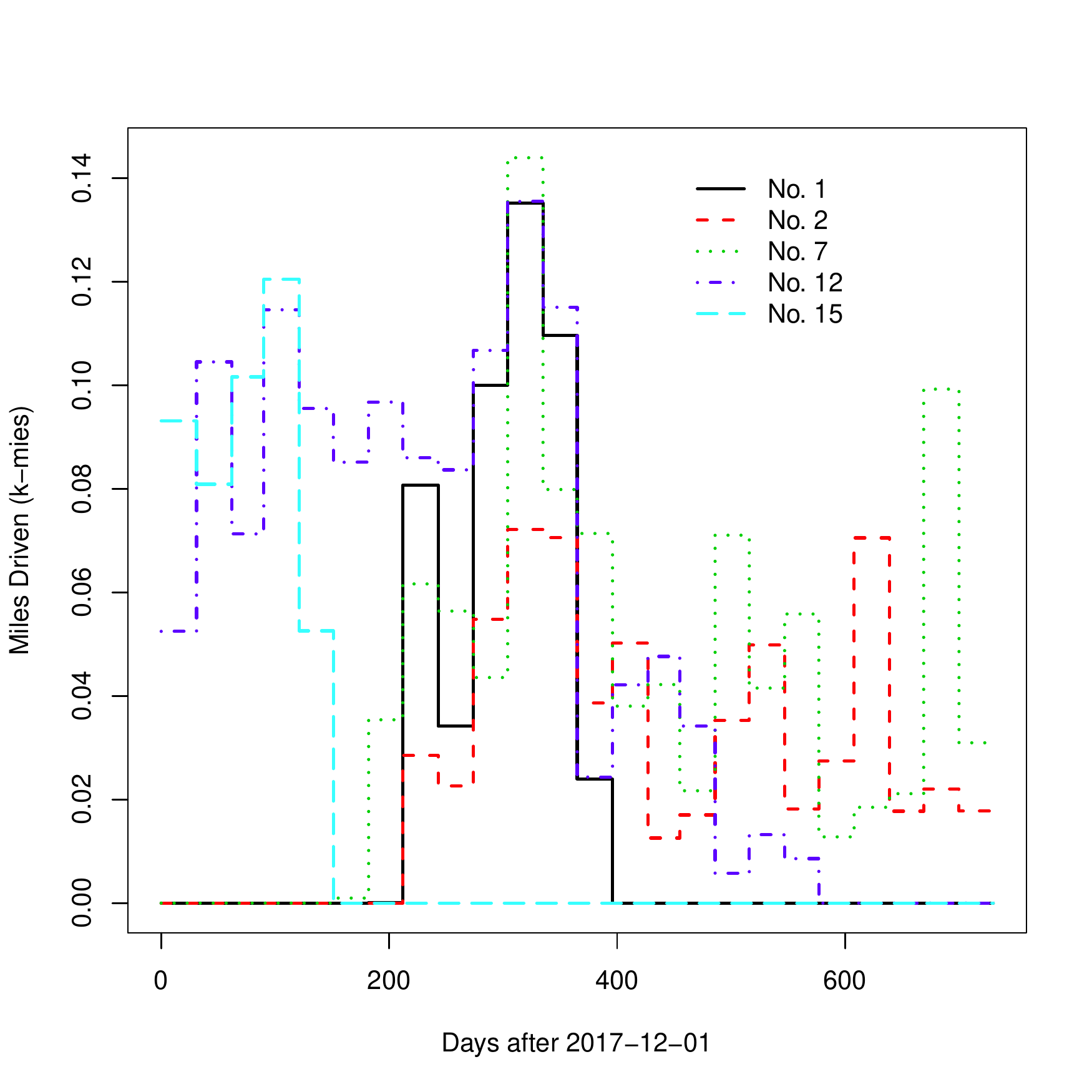}\\
(a) Recurrent Events & (b) Thousands of Miles\\
\end{tabular}
\caption{Visualization of a subset of the recurrent events data and mileage data from manufacturer Waymo. (a) The recurrent events data with the crosses showing the event times and the blue segments showing the active months. (b) A plot of the mileage as a function of time for five representative units. Note that the miles driven are in the units of thousands of miles (k-miles). }\label{fig:sample}
\end{figure}

After data cleaning, the event time is computed as the number of days since the starting date. The monthly mileage is divided by the number of days in that month to obtain the daily mileage, under the approximation that the daily mileage is constant over each month. The unit for the mileage is thousands of miles (k-miles). Figure~\ref{fig:sample} shows a subset of the recurrent events data and mileage data from manufacturer Waymo. Figure~\ref{fig:sample}(a) shows the recurrent events data with the crosses representing the event times and the blue segments showing the active months (i.e., events can only be recorded within the observation windows). Figure~\ref{fig:sample}(b) plots the mileage as a function of time for five representative units. Table~\ref{tab:data.summary} shows a summary of the recurrent events data and mileage data from the four manufacturers. We can see that Waymo and Cruise have driven more than 1 million miles and the disengagement event rate is around 0.1 events per k-miles during the 24 month testing period. Pony AI and Zoox have smaller driving miles, and the event rates are around 0.2 events per k-miles and 0.6 events per k-miles, respectively.

\begin{table}
\caption{Summary of the recurrent events data and mileage data from the four manufacturers over the 24 month study period.}\label{tab:data.summary}
\begin{center}
\begin{tabular}{c|cccccc}\hline\hline
\multirow{2}{*}{Manufacturer}    &  No. of   & Active  & Active Months & No. of & Total in & No. of Events  \\[-0.5ex]
    &  Vehicles & Months & per Vehicle & Events & k-miles & per k-miles \\\hline
Waymo           & 123 & 1550 & 12.602 &  224 & 2710.136 &   0.083 \\\hline
Cruise          & 304 & 2079 & 6.839  &  154 & 1278.661 &   0.120 \\\hline
Pony AI         & 23  & 179  & 7.783  &   43 &  190.871 &   0.225 \\\hline
Zoox            & 32  & 280  & 8.750    & 58 &   97.780 &   0.593 \\\hline\hline
\end{tabular}
\end{center}
\end{table}

\subsection{Notation for Data}
The number of AV testing units (vehicles) in the fleet is denoted by $n$. The total follow-up time is $\tau=730$ days (i.e., two years). Let $t_{ij}, i=1, \ldots, n, j=1, \ldots, n_i$ be event time~$j$ for unit~$i$. Here $t_{ij}$ records the number of days since December 1, 2017, and $n_i$ is the number of events for unit $i$. Note that it is possible that $n_i=0$, indicating that there were no events observed for unit $i$.

Let $x_i(t), 0<t\leq \tau$, be the mileage driven for unit $i$ at time (day) $t$. The unit of $x_i(t)$ is k-miles. Because only monthly mileage was reported, the daily average was used for $x_i(t)$. Thus, $x_i(t)$ is a step function, which can be represented as,
\begin{align}\label{eqn:form.xit}
x_i(t)=\sum_{l=1}^{n_{\tau}}x_{il}\Indfun{(\tau_{l-1}<t\leq\tau_{l})}.
\end{align}
Here, $n_{\tau}=24$ is the number of months in the follow-up period, $x_{il}$ is the daily mileage for unit~$i$ during month $l$, $\tau_{l}$ is the ending day since the starting of the study for month $l$, and $\Indfun{(\cdot)}$ is an indicator function. Let $\xvec_i(t)=\{x_i(s): 0<s\leq t\}$ be the history for the mileage driven for unit $i$.

\section{Statistical Models}\label{sec:stat.model}
\subsection{The Nonhomogeneous Poisson Process}

Recurrent events processes are commonly modeled by a nonhomogeneous Poisson process (NHPP). The event intensity function for unit $i$ is modeled as,
\begin{align}\label{eqn:intensity.function}
\lambda_i[t; x_i(t), \thetavec]=\lambda_0(t; \thetavec)x_i(t).
\end{align}
Here, $\lambda_0(t; \thetavec)=\lambda_0(t)$ is the baseline intensity function (BIF) with parameter vector $\thetavec$. Because $x_i(t)$ is the mileage driven, $\lambda_i[t; x_i(t), \thetavec]$ is the mileage-adjusted event intensity. The BIF can be interpreted as the event rate per k-miles at time $t$ when $x_i(t)=1$. The baseline cumulative intensity function (BCIF) is,
\begin{align}\label{eqn:BCIF}
\Lambda_0(t; \thetavec)=\Lambda_0(t)=\int_{0}^{t}\lambda_0(s; \thetavec)ds.
\end{align}
Note that $\Lambda_0(0; \thetavec)=0$ and $\Lambda_0(t; \thetavec)$ is a non-decreasing function of $t$. The BCIF $\Lambda_0(t)$ can be interpreted as the expected number of events from time 0 to $t$ when $x(t)=1$ for all $t$. The cumulative intensity function (CIF) for unit $i$ is,
\begin{align}\label{eqn:CIF}
\Lambda_i[t; x_i(t), \thetavec]=\int_{0}^{t}\lambda_0(s; \thetavec)x_i(s)ds.
\end{align}

In software reliability, the BCIF and the BIF are used as reliability metrics when recurrent events can be collected from repairable systems (e.g., \citeNP{Wood1996}). The trends in the BIF can indicate the evolution of the reliability in the underlying AV system. For example, improvement of the autonomous technology in AV can lead to a decreasing trend in the BIF (i.e., BCIF increasing at a decreasing rate).

Typically in software reliability, one specifies a parametric form for the BCIF $\Lambda_0(t; \thetavec)$. Note that the BIF can be obtained by taking the derivative of BCIF with respect to $t$. That is, $\lambda_0(t;\thetavec)=d\Lambda_0(t;\thetavec)/dt.$ The commonly used models for $\Lambda_0(t; \thetavec)$ are Musa-Okumoto, Gompertz, and the Weibull models (e.g., \citeNP{Merkel2018}). Table~\ref{tab:para.model.formula} lists their BIFs, BCIFs, and parameters. Note that the Weibull BCIF is similar to the Weibull cumulative distribution function (cdf) but with an asymptote $\theta_1$ as $t$ goes to $\infty$.

\begin{table}
\caption{List of commonly used parametric models and their BIFs, BCIFs, and parameters.}\label{tab:para.model.formula}
\begin{center}
\begin{tabular}{c|c|c|c}\hline\hline
Model    & BCIF $\Lambda_0(t;\thetavec)$ &BIF $\lambda_0(t;\thetavec)$ & Parameters \\\hline
\multirow{2}{*}{Musa-Okumoto} & \multirow{2}{*}{$\theta_1^{-1}\log(1+\theta_2\theta_1 t)$}  & \multirow{2}{*}{$\theta_2(1+\theta_2\theta_1 t)^{-1}$}  &$\thetavec=(\theta_1, \theta_2)'$\\
&&&$\theta_1>0, \theta_2>0$ \\\hline
\multirow{2}{*}{Gompertz} & \multirow{2}{*}{$\theta_1\theta_3^{\theta_2^t}-\theta_1\theta_3$}  &\multirow{2}{*}{$\theta_1\theta_2^t\theta_3^{\theta_2^t}\log(\theta_2)\log(\theta_3)$}  & $\thetavec=(\theta_1, \theta_2, \theta_3)'$\\
&&& $\theta_1>0, 0<\theta_2, \theta_3<1$ \\\hline
\multirow{2}{*}{Weibull}& \multirow{2}{*}{$\theta_1[1-\exp(-\theta_2t^{\theta_3})]$}  & \multirow{2}{*}{$\theta_1\theta_2\theta_3t^{(\theta_3-1)}\exp(-\theta_2t^{\theta_3})$}  & $\thetavec=(\theta_1, \theta_2, \theta_3)'$\\
&&& $\theta_1>0, \theta_2>0, \theta_3>0$ \\\hline\hline
\end{tabular}
\end{center}
\end{table}

\subsection{Spline Models}
Although parametric models can fit certain trends in the event process, they may not be flexible enough to describe the event process for AV testing, as the evolution of the AI technology in an AV system can be complicated, which motivates us to propose the spline model for describing the BCIF. In the spline model, the BCIF is represented as a linear combination of spline bases. That is,
\begin{align}\label{eqn:spline.BCIF}
\Lambda_0(t;\thetavec)=\sum_{l=1}^{n_{s}}\beta_l\gamma_{l}(t), \quad \beta_l\geq 0,\, l=1,\ldots, n_s,
\end{align}
which provides a nonparametric method to describe the BCIF. Here $\thetavec=(\beta_1, \ldots, \beta_{n_s})'$ is the vector for the spline coefficients, $\gamma_{l}(t)$'s are the spline bases, and $n_s$ is the number of spline bases. The BIF can be obtained by taking a derivative with respect to $t$. That is,
\begin{align}\label{eqn:spline.BIF}
\lambda_0(t;\thetavec)=\frac{d\Lambda_0(t;\thetavec)}{dt}=\sum_{l=1}^{n_{s}}\beta_l\frac{d\gamma_{l}(t)}{dt}.
\end{align}

\begin{figure}
\centering
\includegraphics[width=.5\textwidth]{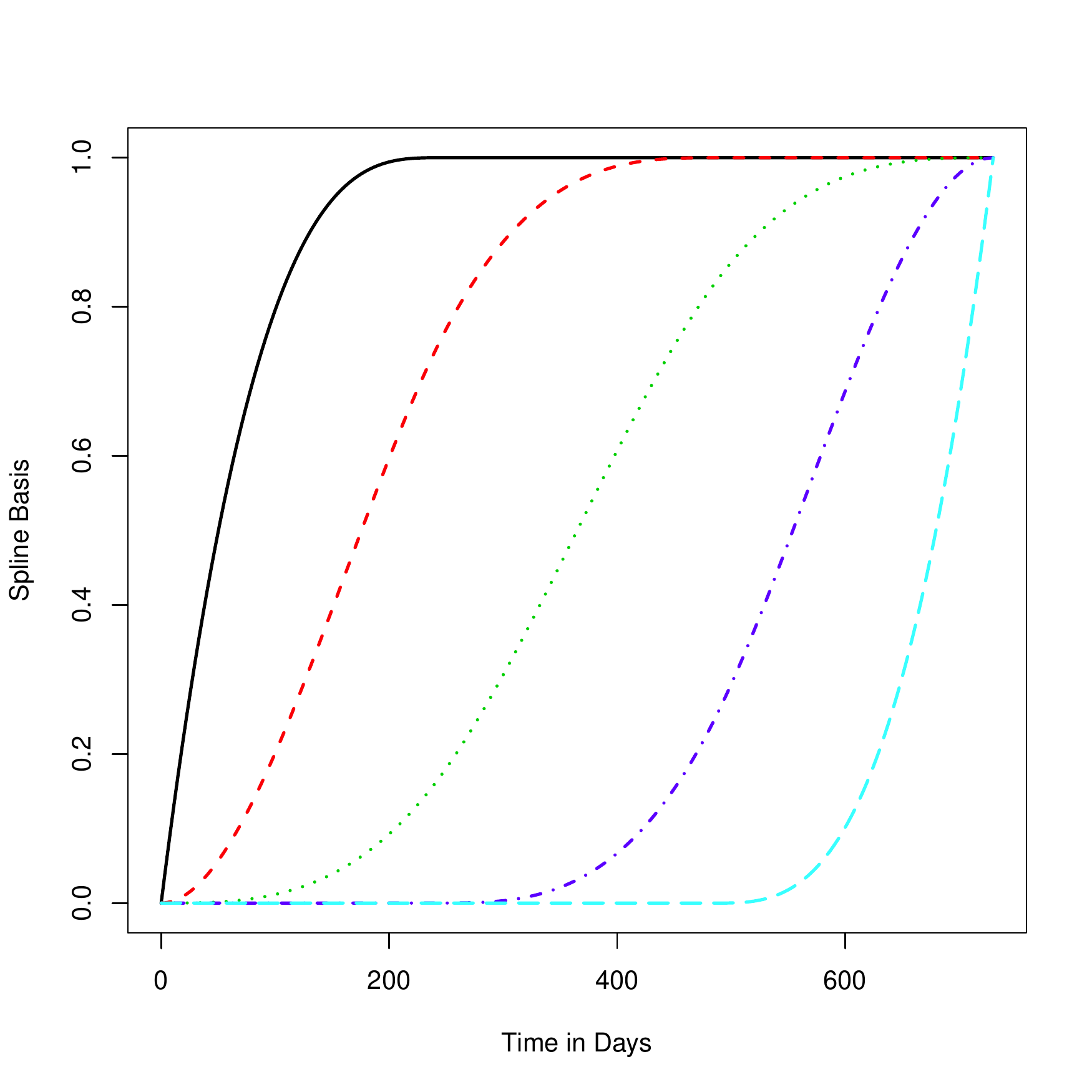}
\caption{Examples of I-spline basis functions.}\label{fig:example.spline.basis}
\end{figure}

Because of the constraints that $\Lambda_0(0; \thetavec)=0$ and that $\Lambda_0(t; \thetavec)$ is a non-decreasing function of $t$, some special considerations are needed in the spline model. We use the I-splines in \citeN{Ramsay1988}. Figure~\ref{fig:example.spline.basis} shows examples of I-spline basis functions (i.e., $\gamma_l(t)$). We can see that each spline basis takes value zero at $t=0$ and is monotonically increasing. By taking non-negative coefficients (i.e., $\beta_l\geq0$), a non-decreasing $\Lambda_0(t; \thetavec)$ is obtained.

A brief introduction on the construction of I-spline basis is as follows. The boundary knots are 0 and $\tau$. The $b$ interior knots are denoted by $t_{h+1},\dots,t_{h+b}$ for splines of order $h$. The complete sequence for the knots are denoted by $0=t_1=\cdots=t_h<t_{h+1}<\cdots<t_{h+b}<t_{h+b+1}=\cdots=t_{2h+b}=\tau.$ The total number of spline bases is $n_s=h+b$. I-splines are obtained by integrating the M-splines; that is, $$I_q^{(h)}(t)=\int_{0}^{t}M_q^{(h)}(u)du, q=1,\dots, b+h, \quad t\in[0, \tau],$$
and the M-spline bases of order $h$ are defined recursively. The M-splines of order 1 is
$$M_q^{(1)}(t)=\Indfun{(t_q\leq z<t_{q+1})}(t_{q+1}-t_q)^{-1},$$
for $q=1,\cdots, b+1$. The M-splines of order $h$ are obtained as
$$M_q^{(h)}(t)=\frac{h[(t-t_q)M_q^{(h-1)}(t)+(t_{q+h}-t)M_{q+1}^{(h-1)}(t)]}{(h-1)(t_{q+h}-t_q)}
\Indfun{(t_q\leq t<t_{q+h})},$$
for $q=1,\cdots, b+h$.

\subsection{Modeling Heterogeneity}
It is maybe possible that some AV units are more likely to generate more events even after accounting for the mileage driven, resulting in heterogeneity in the event process. Similar to the approach in Chapter~3 of \citeN{CookLawless2007}, a frailty term can be added to the parametric intensity function to model this extra heterogeneity. The gamma frailty model is,
\begin{align}\label{eqn:gamma.frailty}
\lambda_i[t; u_i, \xvec_i(t), \thetavec]=u_i\lambda_0(t; \thetavec)x_i(t).
\end{align}
Here, the frailty term $u_i$ is a random variable that follows a gamma distribution with mean one and variance $\phi$. The probability density function (pdf) of $u_i$ is,
\begin{align*}
f(u_i)=\frac{1}{\Gamma(1/\phi)\phi^{1/\phi}}u_{i}^{(1/\phi-1)}\exp(-u_i/\phi).
\end{align*}
The gamma frailty model is popular in reliability and survival applications because there is a closed-form expression for the marginal likelihood for recurrent events data based on the model in~\eqref{eqn:gamma.frailty}.

\section{Model Estimation and Inference}\label{sec:model.est.inf}
This section presents parameter estimation procedures for the parametric and spline models, and the corresponding statistical inference procedures. This section also contains parameter estimation and hypothesis testing for the gamma frailty model.

\subsection{Parameter Estimation}
We use the maximum likelihood (ML) methods for parameter estimation. The likelihood function is,
\begin{align}\label{eqn:event.lik}
L(\thetavec)=\prod_{i=1}^n\left\{\prod_{j=1}^{n_i} \lambda_i[t_{ij}; x_i(t_{ij}), \thetavec]\right\}\times \exp\{-\Lambda_i[\tau; \xvec_i(\tau), \thetavec]\},
\end{align}
with the convention that $\prod_{j=1}^{0} (\cdot)=1$. Here, the intensity function and the CIF are defined in \eqref{eqn:intensity.function} and \eqref{eqn:CIF}, respectively.  The log-likelihood function is obtained by taking the logarithm of the $L(\thetavec)$ in \eqref{eqn:event.lik}. That is,
\begin{align}\label{eqn:log.lik}
l(\thetavec)&=\sum_{i=1}^n\left(\sum_{j=1}^{n_i}\left\{\log[x_i(t_{ij})]+\log[\lambda_0(t_{ij}; \thetavec)]\right\}\right)-\Lambda_i[\tau; \xvec_i(\tau), \thetavec]\\\nonumber
&=\sum_{i=1}^n\sum_{j=1}^{n_i}\left\{\log[x_i(t_{ij})]+\log[\lambda_0(t_{ij}; \thetavec)]\right\}- \sum_{i=1}^{n}\sum_{l=1}^{n_{\tau}}\left\{x_{il}\cdot[\Lambda_0(\tau_{l};\thetavec)-
\Lambda_0(\tau_{l-1};\thetavec)]\right\},
\end{align}
with the convention that $\sum_{j=1}^{0} (\cdot)=0$. The last step in \eqref{eqn:log.lik} is obtained by substituting the expression for $x_i(t)$ in \eqref{eqn:form.xit}.

For parametric models, the functional forms of $\Lambda_0(t)$ and $\lambda_0(t)$ in Table~\ref{tab:para.model.formula} can be substituted into \eqref{eqn:log.lik} to evaluate the log-likelihood function. The ML estimate of $\thetavec$, denoted by $\thetavechat$, can be obtained by finding the value of $\thetavec$ that maximizes $l(\thetavec)$. For parametric models, the dimension of $\thetavec$ is typically 2 or 3. We used the R \texttt{optim()} function with the ``Nelder-Mead'' option to do the optimization.

For the spline model, the functional forms of $\Lambda_0(t)$ and $\lambda_0(t)$ in \eqref{eqn:spline.BCIF} and \eqref{eqn:spline.BIF}, respectively, can be substituted into \eqref{eqn:log.lik} to evaluate the log-likelihood function. To estimate the parameter $\thetavec$ for the spline model, one needs to specify the locations of the knots and the number of knots, and address the non-negativity constraints on $\thetavec$ as indicated in \eqref{eqn:spline.BCIF}.

We use I-splines of degree 3, which is generally smooth enough to fit the BCIF. The boundary knots are set to be 0 on the left side and $\tau=730$ on the right side. The interior knots are set to be equally spaced sample quantiles of the observed event times. For example, if three interior knots are needed, the interior knots are set to the 0.25, 0.5 and 0.75 sample quantiles of the observed event times. After setting the knot locations, the spline bases can be computed.

To select the best number of knots, we use the Akaike information criterion (AIC), which is computed as
$$\textrm{AIC}=-2l(\thetavec)+2df.$$
Here, $df$ is the number of degrees of freedom for the model. For the spline model, $df$ is the number of non-zero coefficients. To address the non-negativity constraints on $\thetavec$, we use the ``L-BFGS-B'' option in the R \texttt{optim()} function. The ``L-BFGS-B'' algorithm allows users to specify an interval for the variable to be optimized. We set the interval to be $[0, \infty)$ for the spline coefficients. Using the ``L-BFGS-B'' algorithm, some of the elements of the estimates $\thetavechat$ could be set to zero.

\subsection{Statistical Inference}
For inference based on parametric models, the normal approximation based on large sample theory can be used. Because the inference for parametric models is relatively straightforward, the focus of this section on the inference for the spline model. For the spline model, the normal approximation is inappropriate because the parameter estimates can occur at the boundary of the parameter space (i.e., $\wh\beta_l$ can be zero). Thus, we use the fractional-random-weight bootstrap (e.g., \shortciteNP{Xuetal2020}). Different from other bootstrap methods, the fractional-random-weight bootstrap provides a convenient way to quantify uncertainty for data with complex structure. For our case, we need to handle recurrent events with window observations and time-varying mileage information, which complicate the structure.

The implementation of the fractional-random-weight bootstrap is straightforward. The log-likelihood function in \eqref{eqn:log.lik} is re-weighted as
\begin{align}\label{eqn:log.lik.wts}
l^{\ast}(\thetavec)=&\sum_{i=1}^{n}\sum_{j=1}^{n_i}w_i\left\{\log[x_i(t_{ij})]+\log[\lambda_0(t_{ij}; \thetavec)]\right\}\\\nonumber
&-\sum_{i=1}^{n}\sum_{l=1}^{n_{\tau}}w_i\left\{x_{il}\cdot[\Lambda_0(\tau_{l};\thetavec)-
\Lambda_0(\tau_{l-1};\thetavec)]\right\},
\end{align}
where the random weights are independently generated from an exponential distribution with mean one. The bootstrap algorithm for generating bootstrap versions of the estimate of BCIF $\wh\Lambda_0^{\ast}(t)$ is summarized as follows.\\[2ex]
\textbf{Algorithm 1: Bootstrap Algorithm for Generating $\wh\Lambda_0^{\ast}(t)$.}
\begin{enumerate}\itemsep 0.0in
\item Generate random weights $w_i, i=1, \ldots, n$, independently from an exponential distribution with mean one.

\item Construct the re-weighted log-likelihood function as in \eqref{eqn:log.lik.wts}.

\item Use AIC to select the best number of knots based on the log-likelihood function in \eqref{eqn:log.lik.wts}.

\item Based on the best number of knots chosen in step 3, the corresponding spline bases, and estimated coefficients, one can compute the estimates for BCIF as $\wh\Lambda_0^{\ast}(t).$

\item Repeat steps 1 to 4 to obtain $B$ copies of $\wh\Lambda_0^{\ast}(t)$, denoted by $\wh\Lambda_0^{\ast b}(t), b=1, \ldots, B$.
\end{enumerate}

Based on the bootstrap estimates $\wh\Lambda_0^{\ast b}(t), b=1, \ldots, B$, one can construct approximate $100(1-\alpha_p)\%$ pointwise confidence interval (PCI) for $\Lambda_0(t)$ for a given $t$,
\begin{align*}
\left[\wh\Lambda_0^{\ast([B\alpha_p/2])}(t),\,\, \wh\Lambda_0^{\ast([B(1-\alpha_p/2)])}(t)\right].
\end{align*}
Here, $\wh\Lambda_0^{\ast(b)}(t)$ is the ordered version of $\wh\Lambda_0^{\ast b}(t)$, and $[\,\cdot\,]$ is the rounding function.

Based on the spline model, one can construct a simultaneous confidence band (SCB) for the BCIF, which can be used to assess the adequacy of the parametric models in fitting the data. An approximate $100(1-\alpha)\%$ SCB for $\Lambda_0(t), t_L\leq t\leq t_U$, is
\begin{align}\label{eqn:scb}
\left[\utilde{\Lambda}_0(t),\,\, \widetilde{\Lambda}_0(t)\right], \quad t_L\leq t\leq t_U,
\end{align}
where $t_L$ and $t_U$ are boundaries to be specified. If an estimated parametric model is contained in the SCB in \eqref{eqn:scb} the parametric model is statistically consistent with the data (i.e., there is no statistical evidence to reject the parametric model). Thus, the SCB constructed by the spline model provides a tool to check if a parametric model is adequate.

The $100(1-\alpha_p)\%$ PCIs provide a structure for computing SCBs for $\Lambda_0(t)$ for all in the $t\in [t_L,\, t_U]$ period. We use bootstrap samples to calibrate the PCIs so that it can approximately provide the nominal $100(1-\alpha)\%$ CP, similar to the idea of the equal-precision SCB for a cdf in \citeN{Nair1984}. The CP can be estimated as
\begin{align*}
\textrm{CP}(\alpha_p)=\frac{1}{B}\sum_{b=1}^{B}\Indfun{\left(\wh\Lambda_0^{\ast([B\alpha_p/2])}(t)\leq \wh\Lambda_0^{\ast b}(t)\leq \wh\Lambda_0^{\ast([B(1-\alpha_p/2)])}(t),\,\,\textrm{for all } t\in [t_L,\, t_U]\right)}.
\end{align*}
By setting $\textrm{CP}(\alpha_p)=1-\alpha$, one can find the solution to be $\alpha_c$. Thus, the SCB in \eqref{eqn:scb} can be computed as
\begin{align*}
\left[\wh\Lambda_0^{\ast([B\alpha_c/2])}(t),\,\, \wh\Lambda_0^{\ast([B(1-\alpha_c/2)])}(t)\right], \quad t_L\leq t\leq t_U,
\end{align*}
which is time-efficient because the bootstrap samples have already been obtained.

\subsection{The Frailty Model}\label{sec:frailty}
To estimate the gamma frailty in \eqref{eqn:gamma.frailty}, one needs to calculate the marginal likelihood function. The marginal likelihood function is,
\begin{align}
L(\thetavec, \phi)&=\prod_{i=1}^n \int_{0}^{\infty}\left\{\prod_{j=1}^{n_i} u_i\lambda_i[t_{ij}; x_i(t_{ij}), \thetavec]\right\}\times \exp\{-u_i\Lambda_i[\tau; \xvec_i(\tau), \thetavec]\}f(u_i)du_i\\\nonumber
&=\prod_{i=1}^n\left\{\prod_{j=1}^{n_i} \lambda_i[t_{ij}; x_i(t_{ij}), \thetavec]\right\}\times\int_{0}^{\infty}u_i^{n_i}\exp(-u_ic_i)f(u_i)du_i\\\nonumber
&=\prod_{i=1}^n\left\{\prod_{j=1}^{n_i} \lambda_i[t_{ij}; x_i(t_{ij}), \thetavec]\right\}\times\frac{\phi^{-1/\phi}\Gamma(n_i+1/\phi)}{\Gamma(1/\phi)(c_i+1/\phi)^{(n_i+1/\phi)}}\,,
\end{align}
where $c_i=\sum_{l=1}^{n_{\tau}}x_{il}\cdot[\Lambda_0(\tau_{l};\thetavec)-
\Lambda_0(\tau_{l-1};\thetavec)]$. The ML estimates of $\thetavec$ and $\phi$ are obtained by finding the values that maximize the log-likelihood function, $\log[L(\thetavec, \phi)]$. We again use the R \texttt{optim()} function with the ``Nelder-Mead'' option to do the optimization.

To check if there is population heterogeneity in the event process, one can use the likelihood ratio test. The test statistic is constructed as,
\begin{align}
-2\{l(\wh\thetavec)-\log[L(\wh\thetavec, \wh\phi)]\},
\end{align}
which has a $\chi_1^2$ distribution under the null hypothesis that $\phi=0$. Here, $l(\wh\thetavec)$ is obtained by evaluating \eqref{eqn:log.lik} at $\thetavechat$. The null hypothesis is rejected if the test statistic is larger than $\chi_{1,\,(1-\alpha)}^2$, indicating that there is population heterogeneity.

\section{Simulation Study}\label{sec:simulation.study}
The purpose of the simulation study is to show the properties of the ML estimator for the spline model, to check the CP of the SCB procedure based on the spline model, and see if the SCB can correctly accept or reject a particular parametric model.

\subsection{Setting}
\begin{figure}
\centering
\includegraphics[width=.5\textwidth]{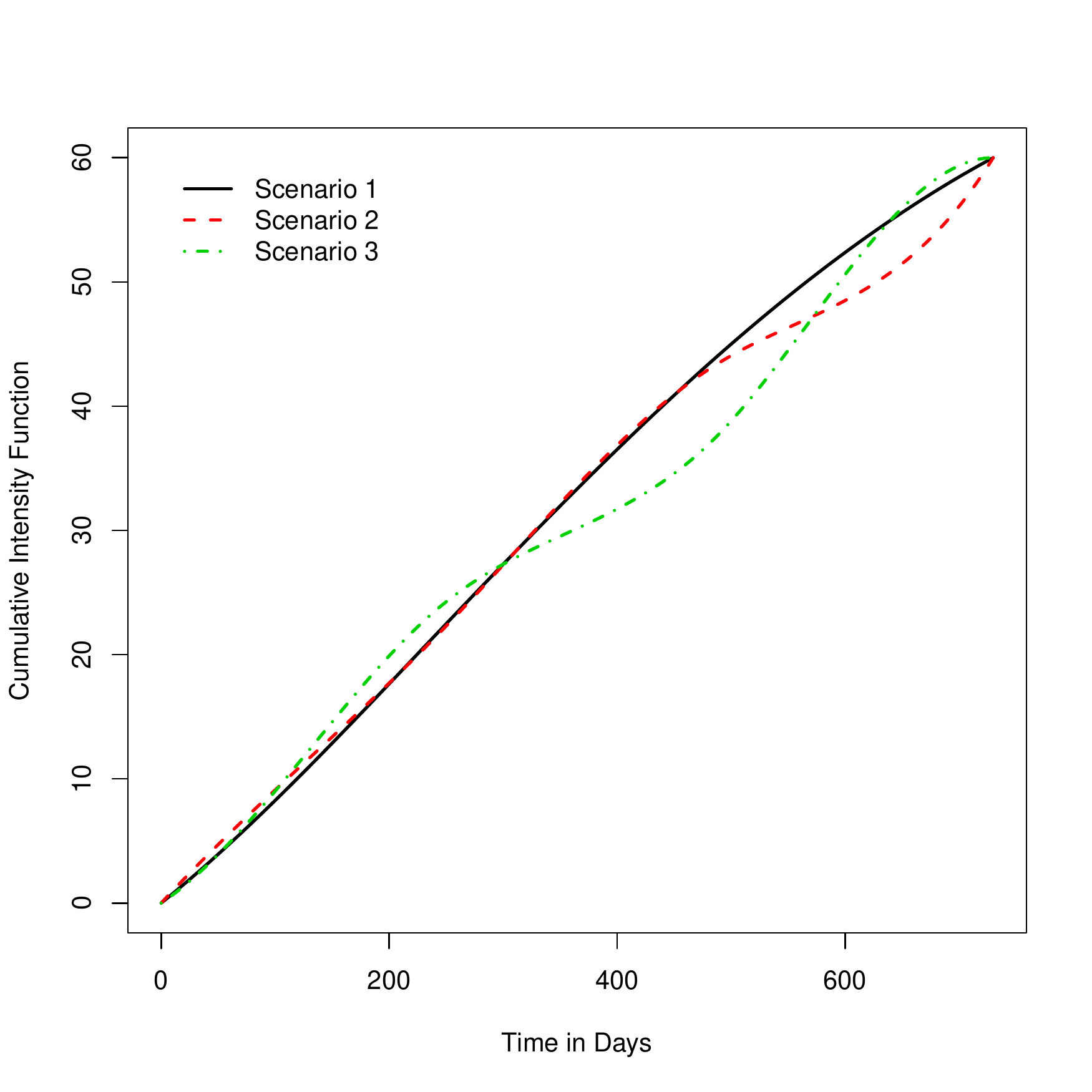}
\caption{Plot of the true BCIFs used in the three simulation scenarios.}\label{fig:sim.true.funs}
\end{figure}

In the simulation study, the spline model is considered as the true underlying model. The spline bases are shown in Figure~\ref{fig:example.spline.basis}. We consider three scenarios. Figure~\ref{fig:sim.true.funs} shows the true BCIFs used in the three simulation scenarios. To simplify the setting, we only consider one parametric model, which is the Gompertz model.

\begin{itemize}
\item Scenario 1:  we choose $\thetavec=(6, 16, 23, 11, 4)'$ as the coefficients. Using the spline bases in Figure~\ref{fig:example.spline.basis}, the true BCIF is shown as the black/solid line in Figure~\ref{fig:sim.true.funs}. The Gompertz model fits this BCIF perfectly.

\item Scenario 2: we choose $\thetavec=(8, 12, 28, 0, 12)'$ as the coefficients. The corresponding true BCIF is shown as the red/dash line in Figure~\ref{fig:sim.true.funs}. The Gompertz model fits this BCIF well except for the late stage.

\item Scenario 3: we choose $\thetavec=(5, 25, 0, 30, 0)'$ as the coefficients. The corresponding true BCIF is shown as the green/dot-dash line in Figure~\ref{fig:sim.true.funs}. The Gompertz model does not fit this BCIF well due to various changes in slope over time.
\end{itemize}

The number of bootstrap samples is $B=5000$ and the number of simulated datasets (i.e., repeats) is $N=1000$. The mileage driven history is sampled with replacement from the historical data from Waymo. The average number of events per unit is around 1.8. The sample sizes (i.e., the number of AV units) considered in the simulation are 50, 100, 200, 500, and 1000.

We evaluate the relative root mean squared errors (RMSE) for the BCIF estimator, the CP for the SCB, and the acceptance probability for the parametric model. In particular, the relative RMSE (RelRMSE) is computed as
\begin{align*}
\textrm{RelRMSE}=\frac{\left\{\sum_{l=1}^N\left[\wh\Lambda_{0l}(t)-\Lambda_0(t)\right]^2/N\right\}^{1/2}}
{\Lambda_0(t)},
\end{align*}
where $\wh\Lambda_{0l}(t)$ is the estimated BCIF using the spline model based on the $l$th simulated dataset, and $\Lambda_0(t)$ is the true BCIF. We use the relative RMSE to remove the scale effect of $\Lambda_0(t)$. That is, RMSE tends to be large if $\Lambda_0(t)$ is large at a particular $t$. The CP is estimated by the proportion that the SCB constructed by using the spline model captures the true BCIF. The acceptance probability is estimated by the proportion of times that the spline-based SCB captures the estimated BCIF from the Gompertz model.

\subsection{Results}
\begin{figure}
\centering
\begin{tabular}{cc}
\includegraphics[width=.4\textwidth]{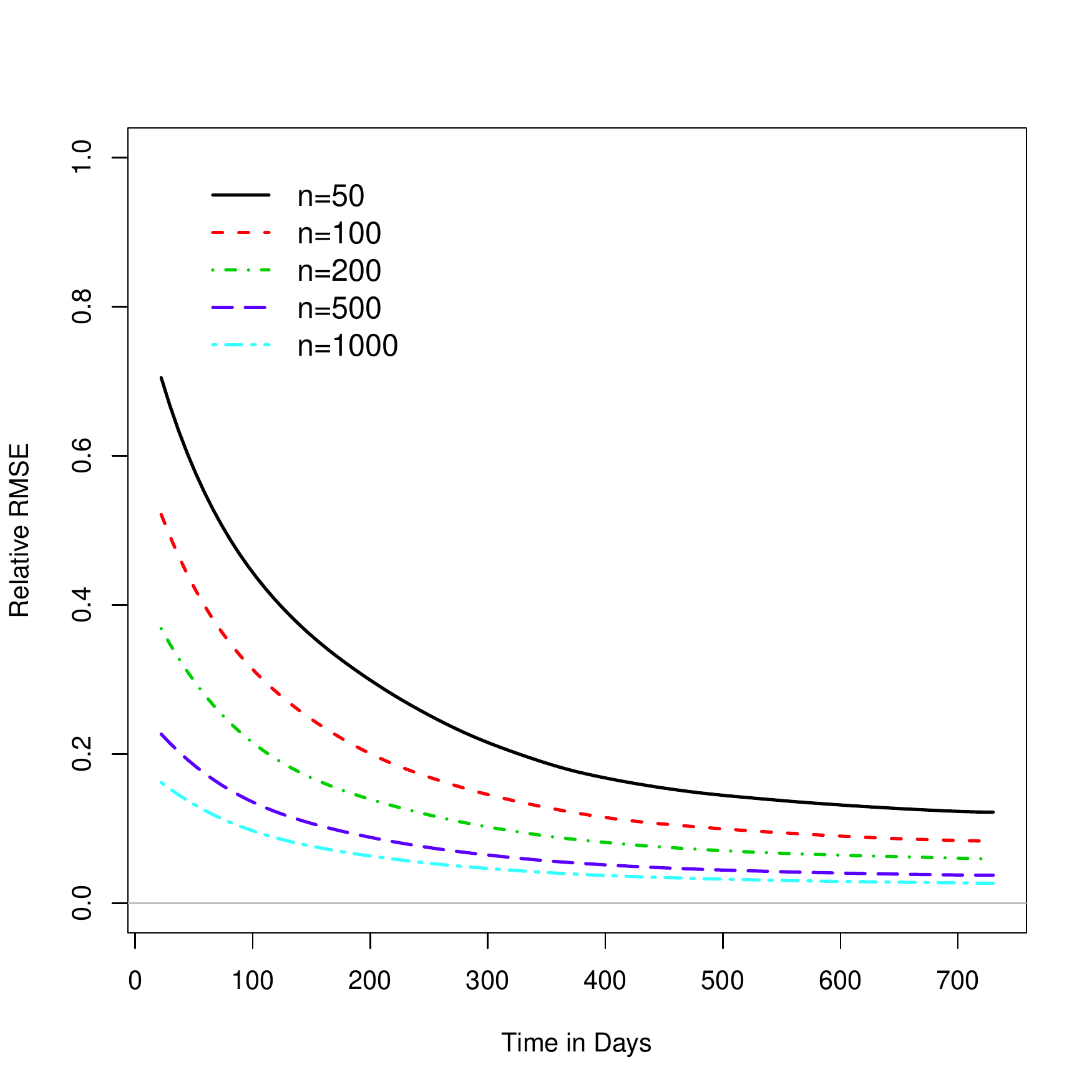}&
\includegraphics[width=.4\textwidth]{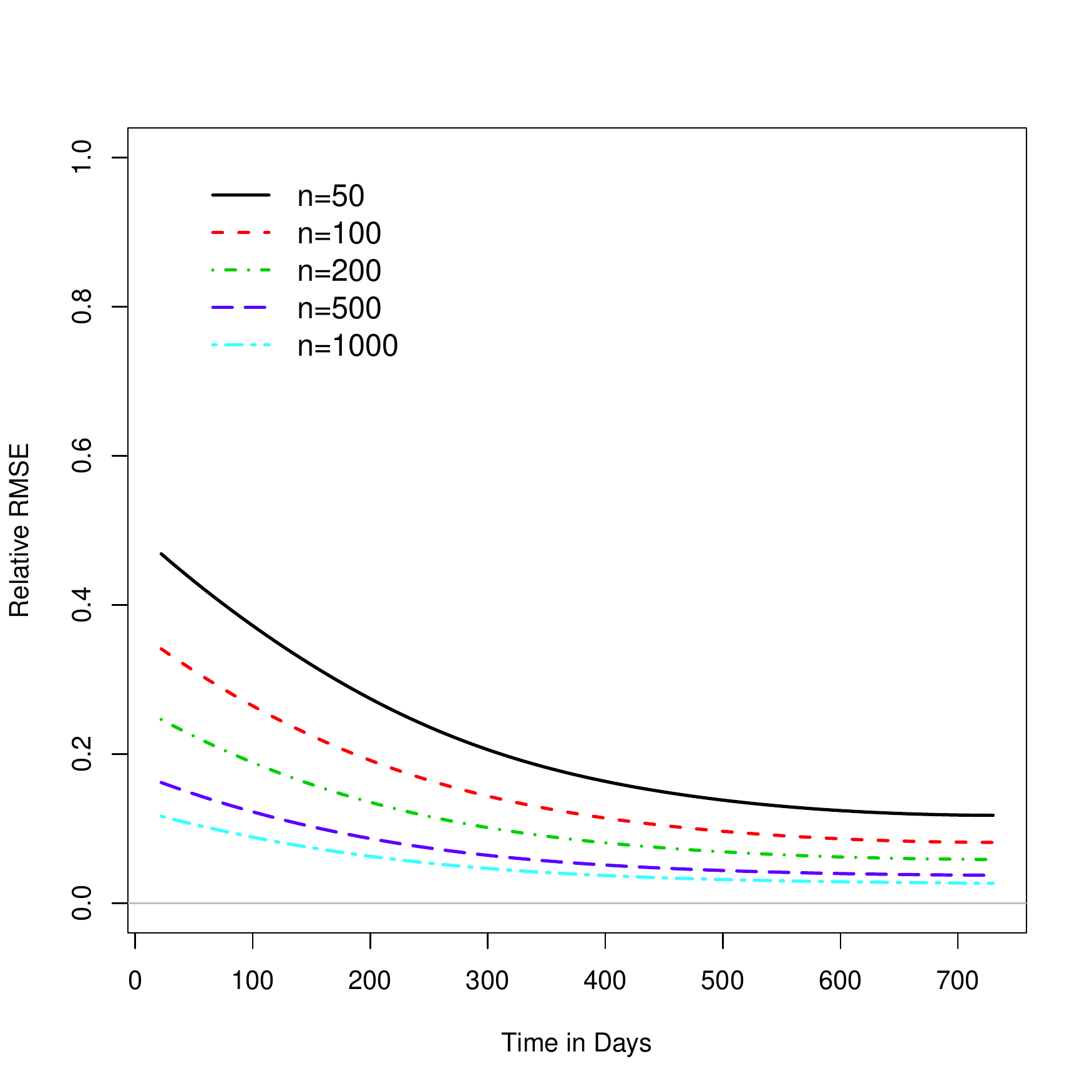}\\
(a) Scenario 1, Spline Model & (b) Scenario 1, Gompertz\\
\includegraphics[width=.4\textwidth]{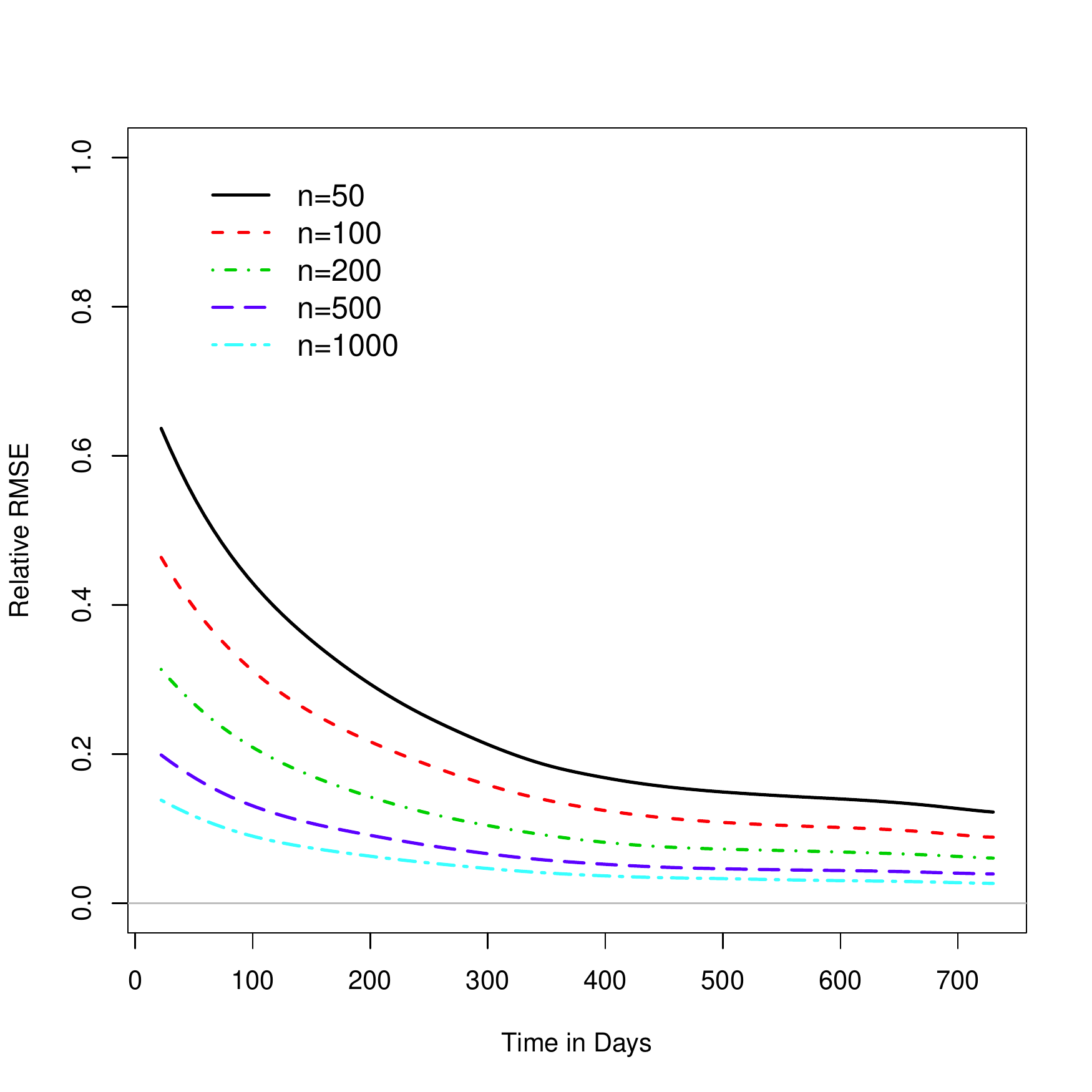}&
\includegraphics[width=.4\textwidth]{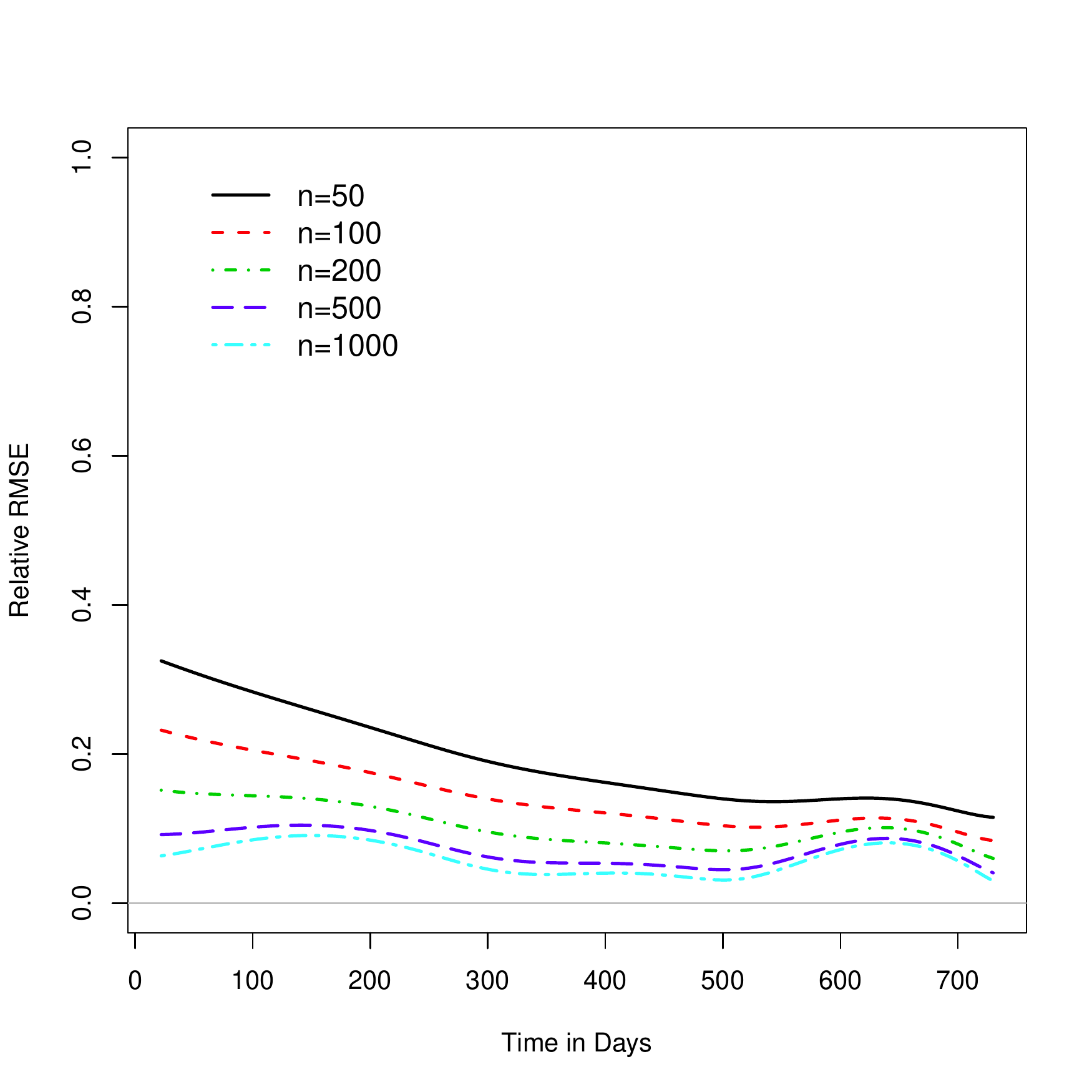}\\
(c) Scenario 2, Spline Model & (d) Scenario 2, Gompertz\\
\includegraphics[width=.4\textwidth]{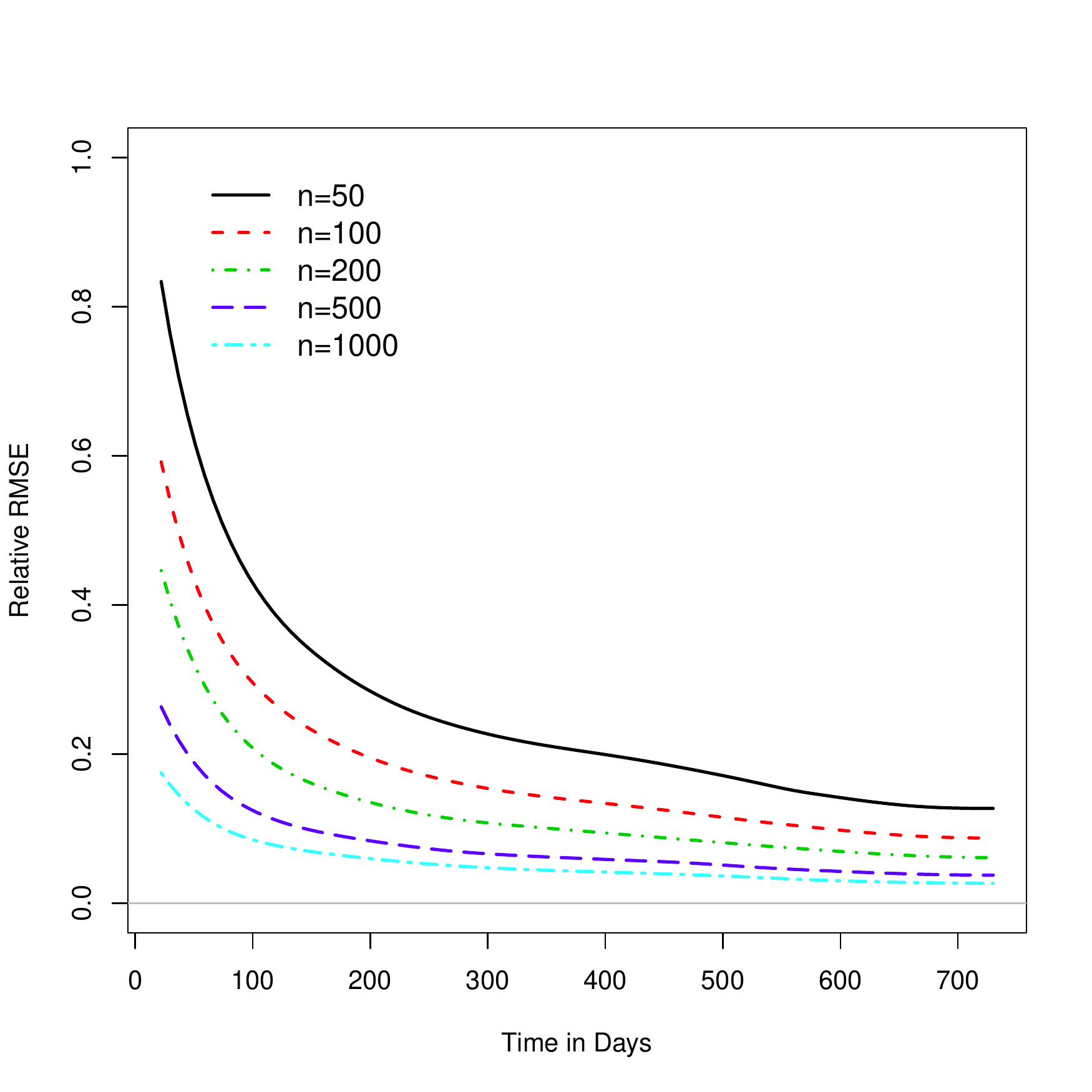}&
\includegraphics[width=.4\textwidth]{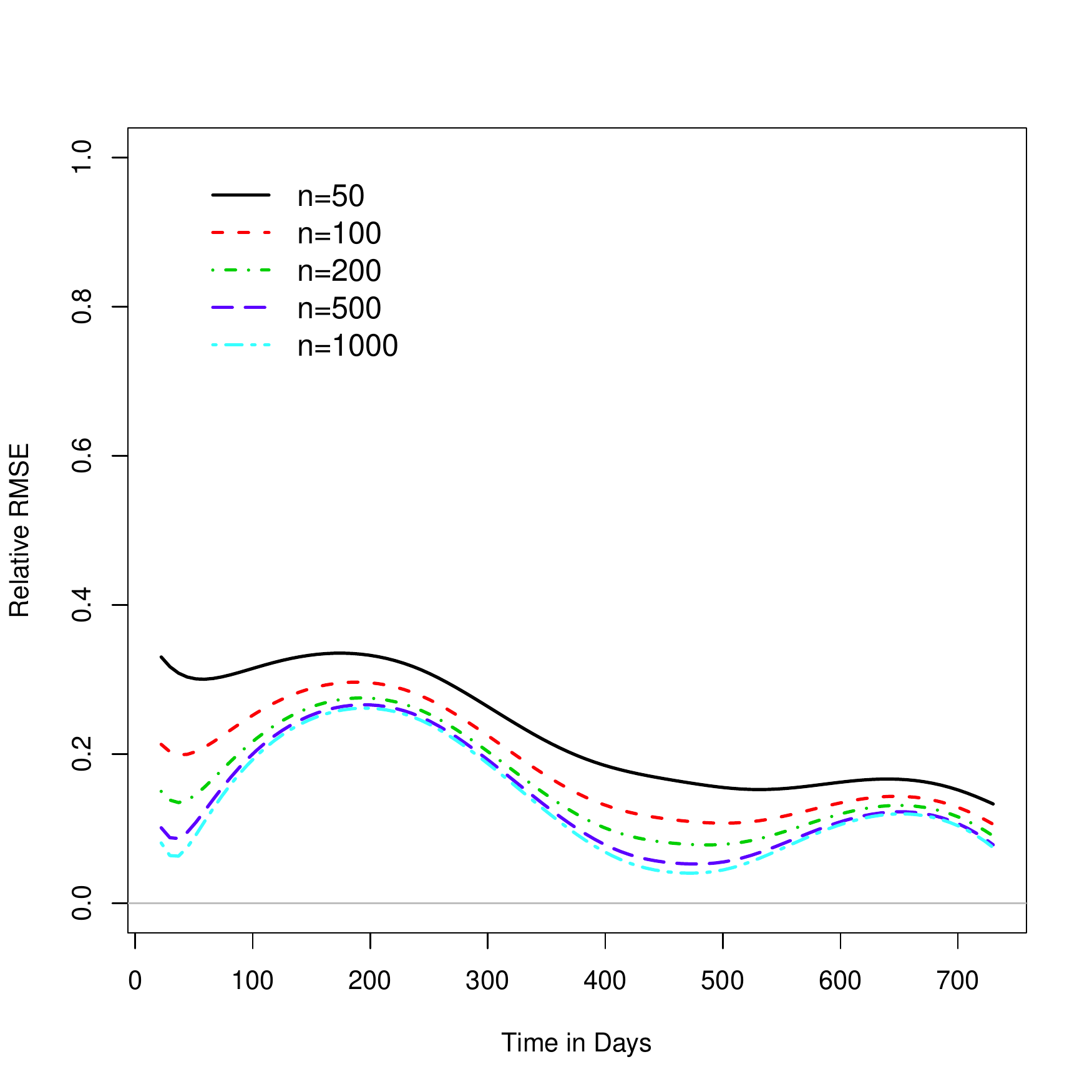}\\
(e) Scenario 3, Spline Model & (f) Scenario 3, Gompertz\\
\end{tabular}
\caption{Plots of relative RMSE as a function of time using the spline model and Gompertz model to fit the data under the three scenarios.}\label{fig:rmse}
\end{figure}

Figure~\ref{fig:rmse} shows the plots of the RelRMSE as a function of time using the spline model and the Gompertz model to fit the data under the three scenarios. As we can see from the figure, the RelRMSE is generally decreasing as the sample size increases for the spline model across all the three scenarios. When the sample size is large, the RelRMSE for the spline model estimator is within a small range.

The behavior of the RelRMSE for the Gompertz model depends on the scenario. For Scenario~1, in which the Gompertz model fits well to the true model, the spline model tends to have a higher RelRMSE because there are more parameters in the spline model (and thus more variability in the estimates) than in the parametric model. For Scenario~2, in which there is some departure in the late stage, the Gompertz model has smaller RelRMSE than the spline model but the advantage diminishes when the sample size increases because bias begins to dominate variance. For Scenario 3, in which there is a large difference from the true model, the Gompertz model tends to have larger RelRMSE than the spline model and the RelRMSE does not decrease much when the sample size increases due to the effect of large bias.

In summary, the ML estimator for the spline model works as expected.  When a parametric BCIF is adequate, it tends to have higher statistical efficiency. When the parametric model is not adequate, there could be large RelRMSE due to bias in the estimation. The results show that it is important to assess the adequacy of parametric models. The spline model, however, is flexible at the price of losing some statistical efficiency.

\begin{figure}
\centering
\begin{tabular}{cc}
\includegraphics[width=.48\textwidth]{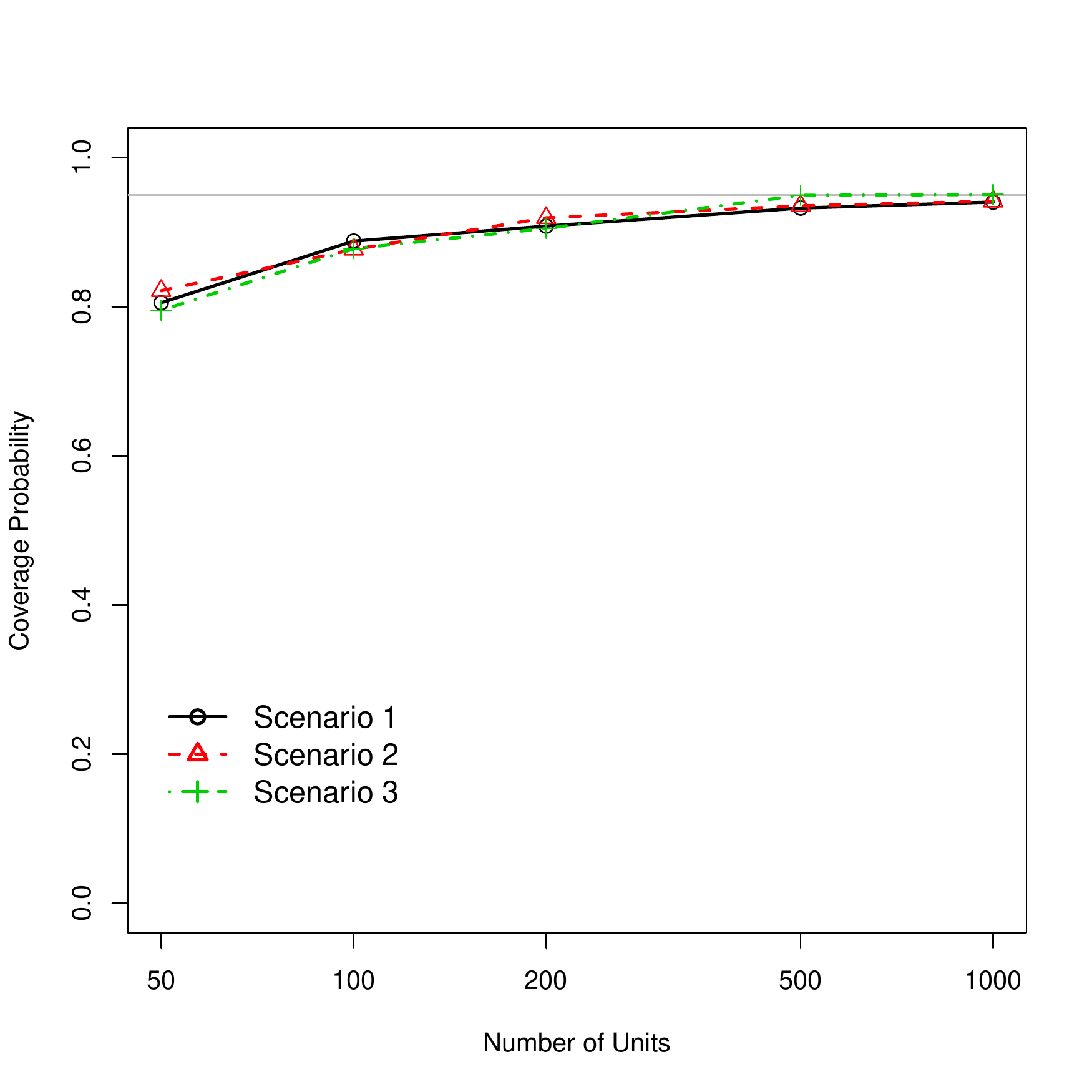}&
\includegraphics[width=.48\textwidth]{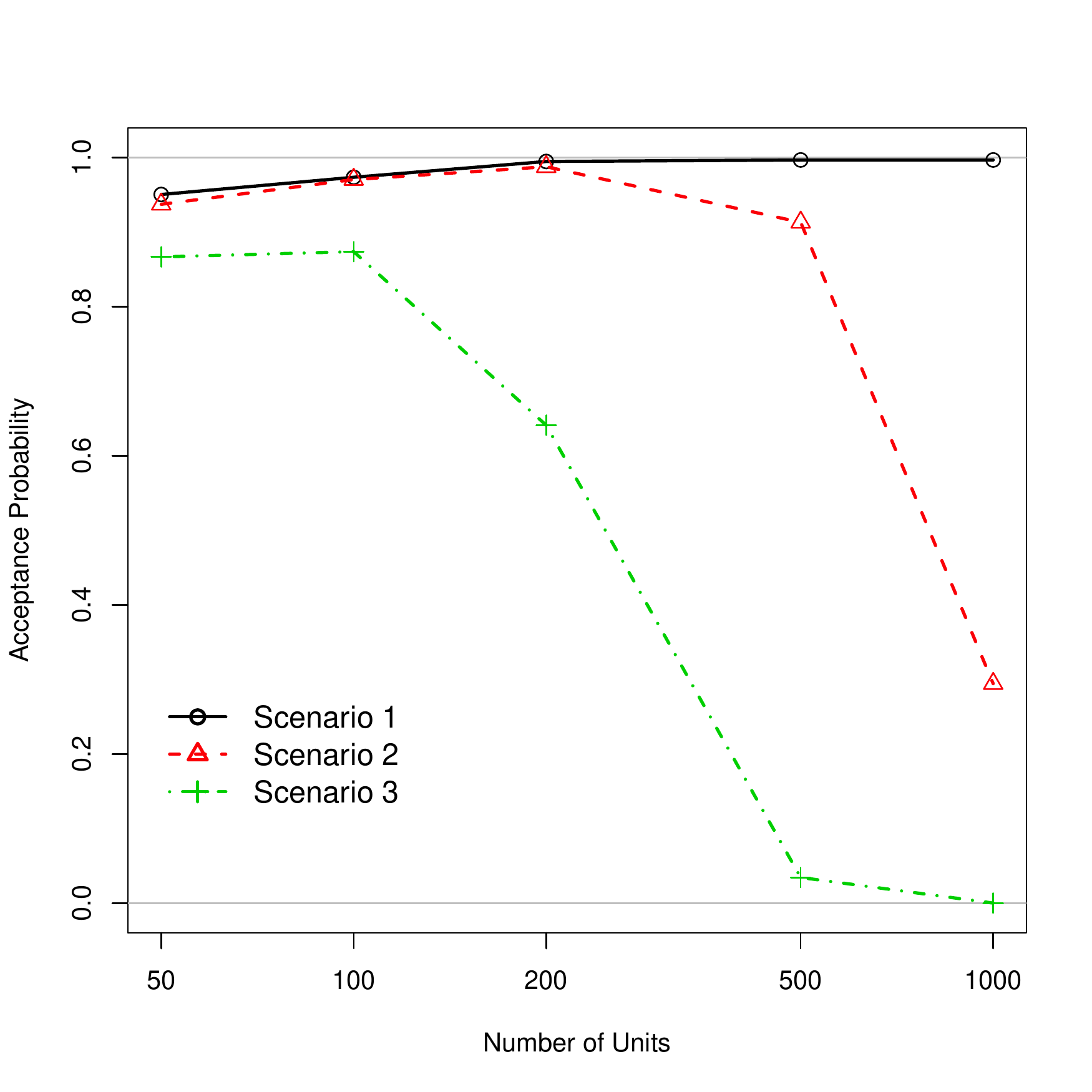}\\
(a) Coverage Probability & (b) Acceptance Probability\\
\end{tabular}
\caption{Plots of coverage probability and acceptance probability as a function of sample size under the three scenarios.}\label{fig:cp.accept}
\end{figure}

Figure~\ref{fig:cp.accept} shows the plot of CP and acceptance probability as a function of sample size under the three scenarios. From Figure~\ref{fig:cp.accept}(a), the CP for the SCB procedure based on the spline model and bootstrap is quite similar for the three scenarios. The CP improves when the sample size increases. The CP is less than nominal when the sample size is small and is getting closer to the nominal CP when the sample is larger than 200. From the plot of the acceptance probability in Figure~\ref{fig:cp.accept}(b), the parametric model is generally accepted when the sample size is small and there is  little or no departure from the true model. When there is a large departure from the true model, as in Scenario 3, the SCB does not capture the estimated Gompertz model with high probability when the sample size is large.

\section{Data Analysis}\label{sec:data.analysis}
In this section, we present the data analysis for the recurrent disengagement events data.

\subsection{Model Fitting}
We start by fitting the parametric models in Table~\ref{tab:para.model.formula} and the spline model to the disengagement-events data from the four manufacturers: Waymo, Cruise, Pony AI, and Zoox. The model fitting uses the ML estimation procedures described in Section~\ref{sec:model.est.inf}. Table~\ref{tab:aic.tab} shows AIC values for the selected models fitted to the data from the four manufacturers.  The numbers with bold font indicate the lowest AIC values among the parametric models. The spline model, in general, results in much lower AIC values except for Zoox data. This shows that the spline model is flexible in fitting the recurrent event data. For Zoox, the AIC value for the Musa-Okumoto model is small because the model has two parameters. Based on the AIC values, the best parametric model for Waymo and Cruise is the Gompertz model. The best parametric model for Pony AI is the Weibull model, and the best parametric model for Zoox is the Musa-Okumoto model.

\begin{table}
\caption{The values of AIC for fitting various models to the data from the four manufacturers. The numbers with bold font indicate the lowest AIC values among the parametric models.}\label{tab:aic.tab}
\begin{center}
\begin{tabular}{c|ccccc}\hline\hline
\multirow{2}{*}{Manufacturer} &&\multicolumn{3}{c}{Parametric Models} &\multirow{2}{*}{Spline Model} \\\cline{3-5}
       &&     Musa-Okumoto  &    Gompertz  &    Weibull  &      \\\hline
Waymo  && 2769.78 & \textbf{2769.70} & 2770.60  & 2756.21  \\\hline
Cruise && 2051.27 & \textbf{2047.42} & 2048.28  & 2046.09  \\\hline
Pony AI && 499.79  & 504.56  & \textbf{498.73}   &  479.73  \\\hline
Zoox    && \textbf{687.69}  & 689.55 & 689.38    & 688.78   \\\hline\hline
\end{tabular}
\end{center}
\end{table}

\begin{figure}
\centering
\begin{tabular}{cc}
\includegraphics[width=.48\textwidth]{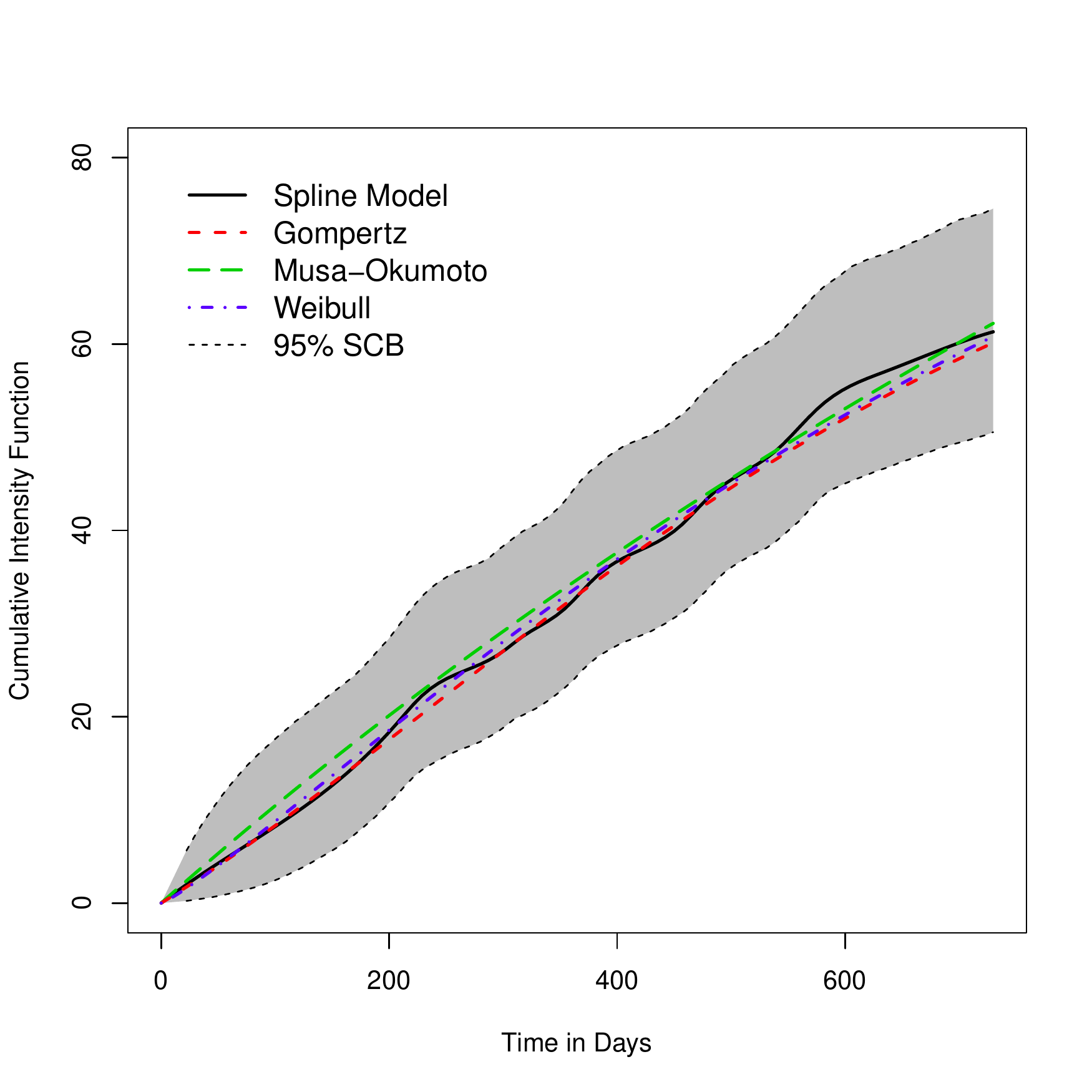}&
\includegraphics[width=.48\textwidth]{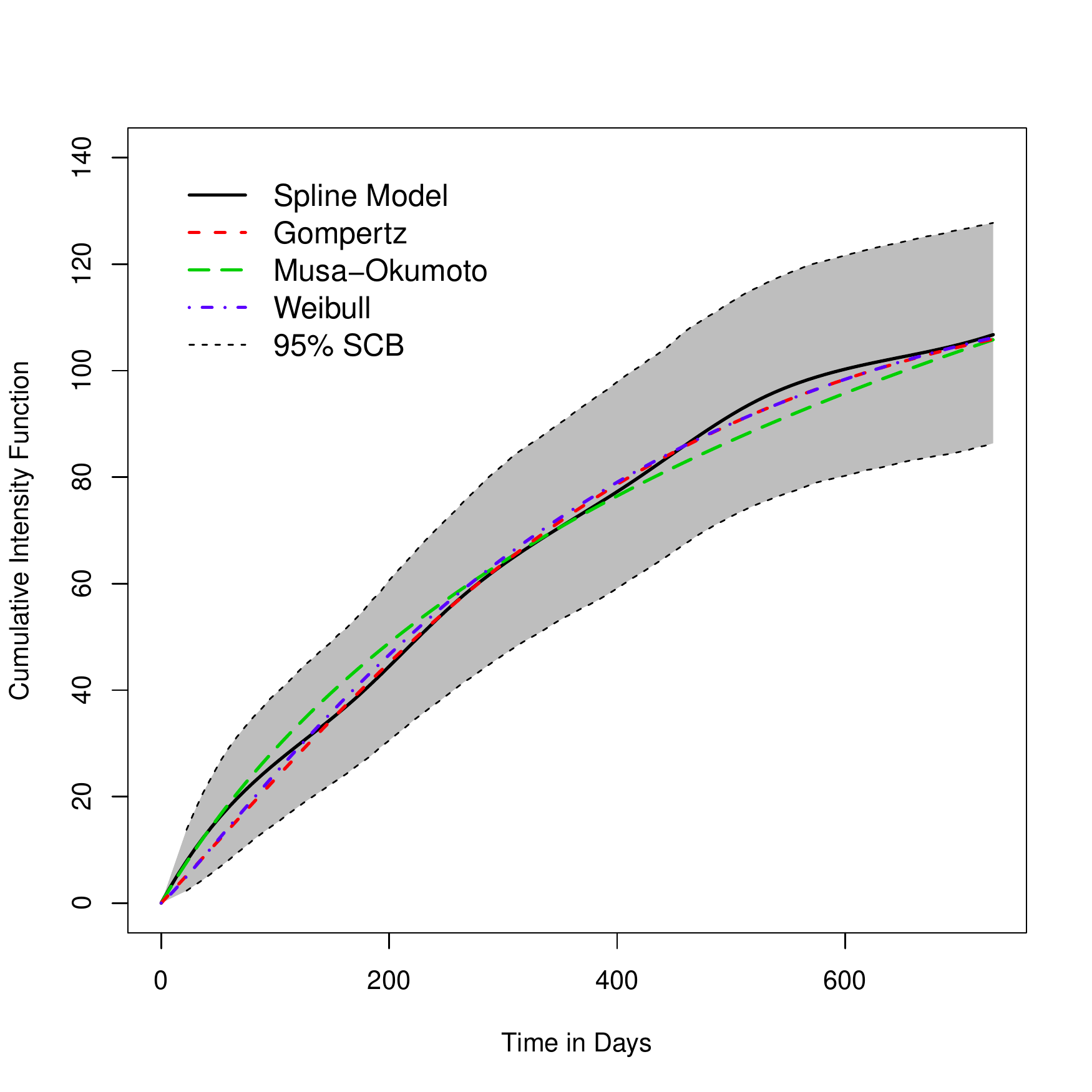}\\
(a) Waymo & (b) Cruise\\
\includegraphics[width=.48\textwidth]{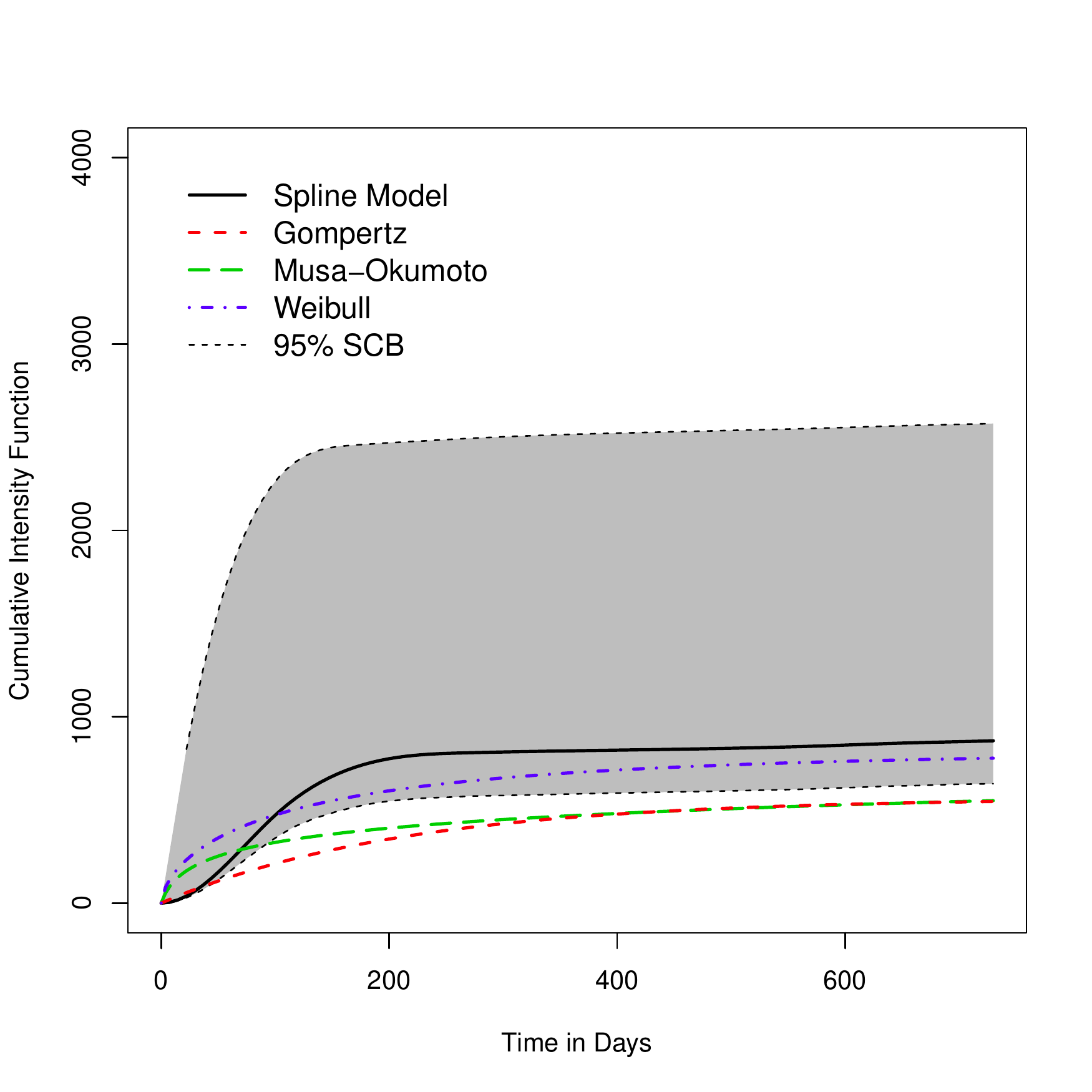}&
\includegraphics[width=.48\textwidth]{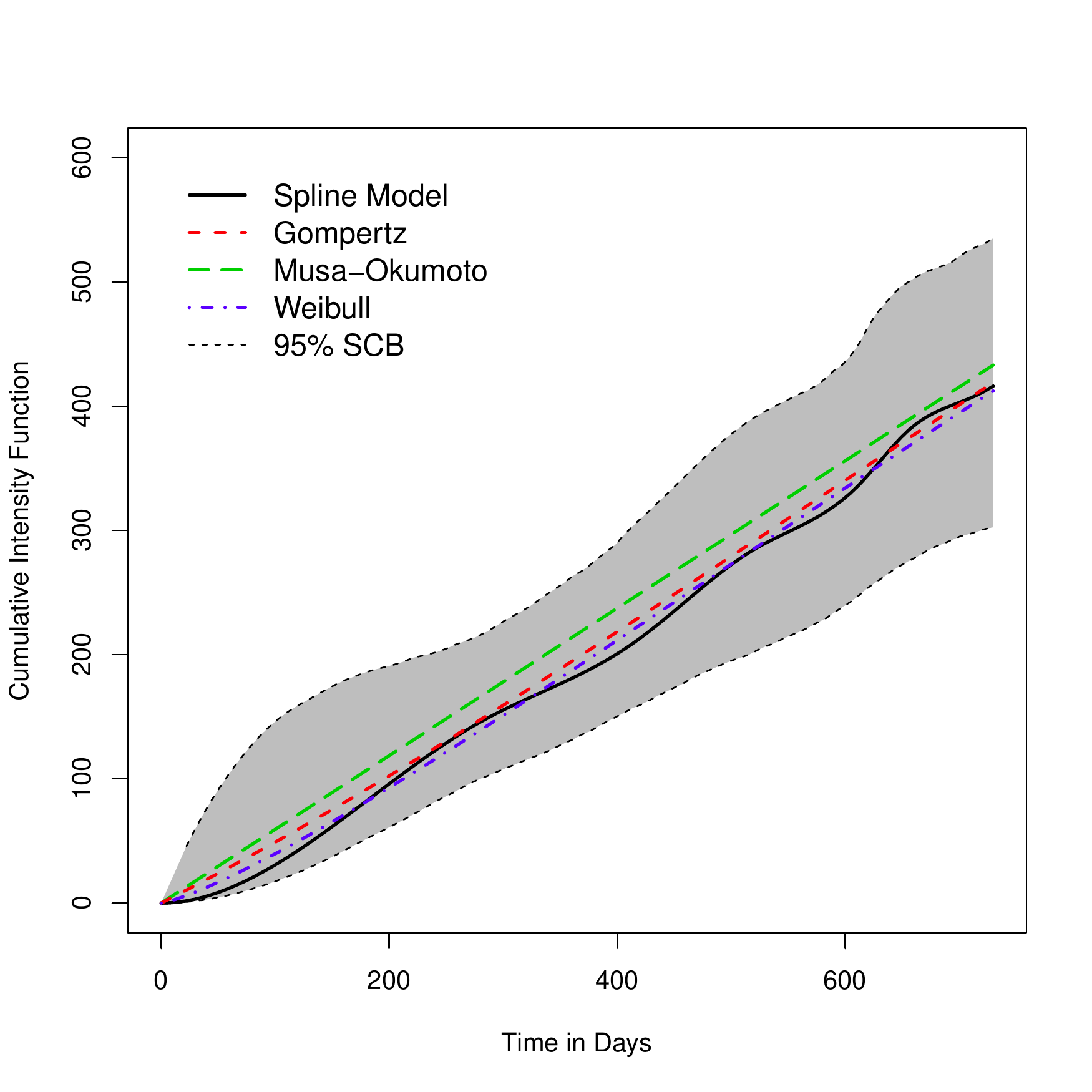}\\
(c) Pony AI & (d) Zoox\\
\end{tabular}
\caption{Plots of the estimated BCIFs for the four manufacturers based on the spline model and parametric models, together with the 95\% SCB based on the spline model.}\label{fig:cum.int.fun}
\end{figure}

To visualize the model estimation results, Figure~\ref{fig:cum.int.fun} shows the plots of the estimated BCIFs for the four manufacturers based on the spline model and parametric models, together with the 95\% SCB based on the spline model.  For Waymo and Cruise, all the parametric models are within the SCB and agree quite well with the estimated BCIF from the spline model. The Gompertz model agrees with the observed number of events better than other models over certain time ranges.

For Pony AI, the SCB is wide, indicating large variability in the estimation. In the early stage of the testing (i.e., from day 0 to day 200), all the events were coming from two units with a lower mileage driven at $x_i(t)=0.01$. A high number of events with a low mileage driven leads to a high event rate. The SCB is asymmetric because it was built using fractional-random-bootstrap and the distribution of $\wh\Lambda_0(t)$ is heavily skewed.  The fact that all events come from two units leads to a large amount of variability in the SCB as the change of weights of the two units has a large influence on the re-weighted log-likelihood.  We also note that the best parametric model (the Weibull) does not agree well with the spline model, although it tracks the trend. For Zoox, the SCB is also relatively wide due to the small number of test units. All parametric models are within the SCB, indicating all parametric models are statistically acceptable.


\begin{figure}
\centering
\begin{tabular}{cc}
\includegraphics[width=.48\textwidth]{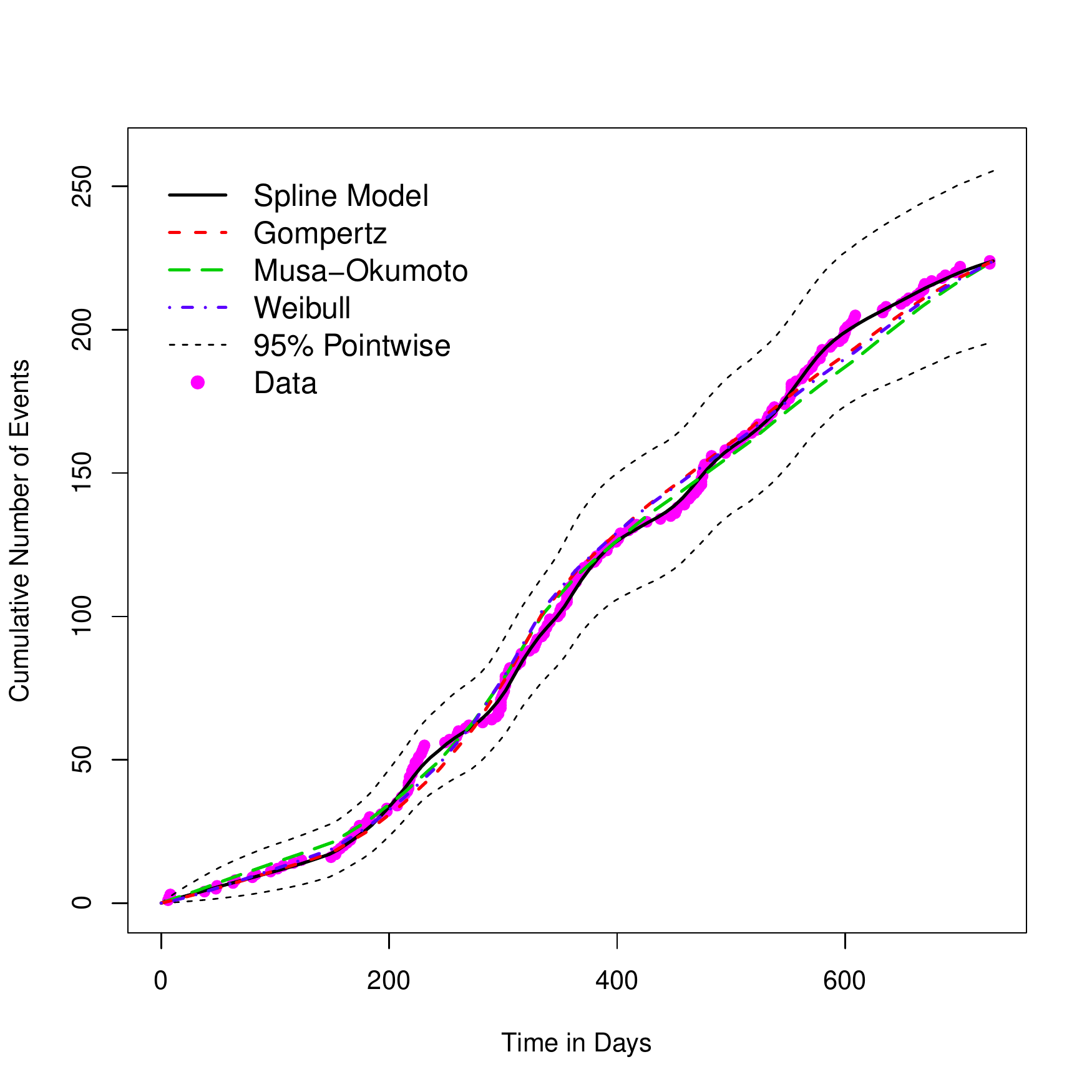}&
\includegraphics[width=.48\textwidth]{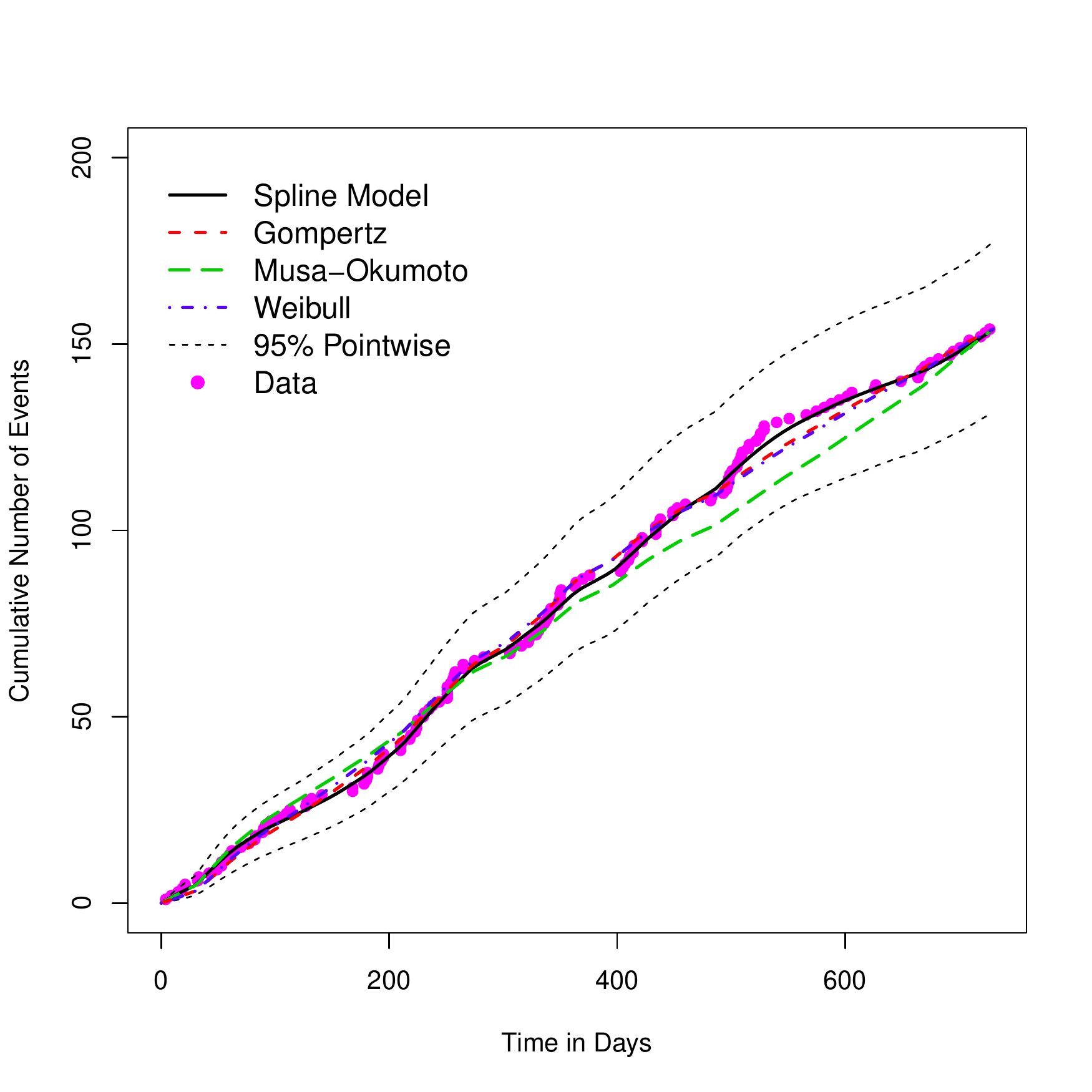}\\
(a) Waymo & (b) Cruise\\
\includegraphics[width=.48\textwidth]{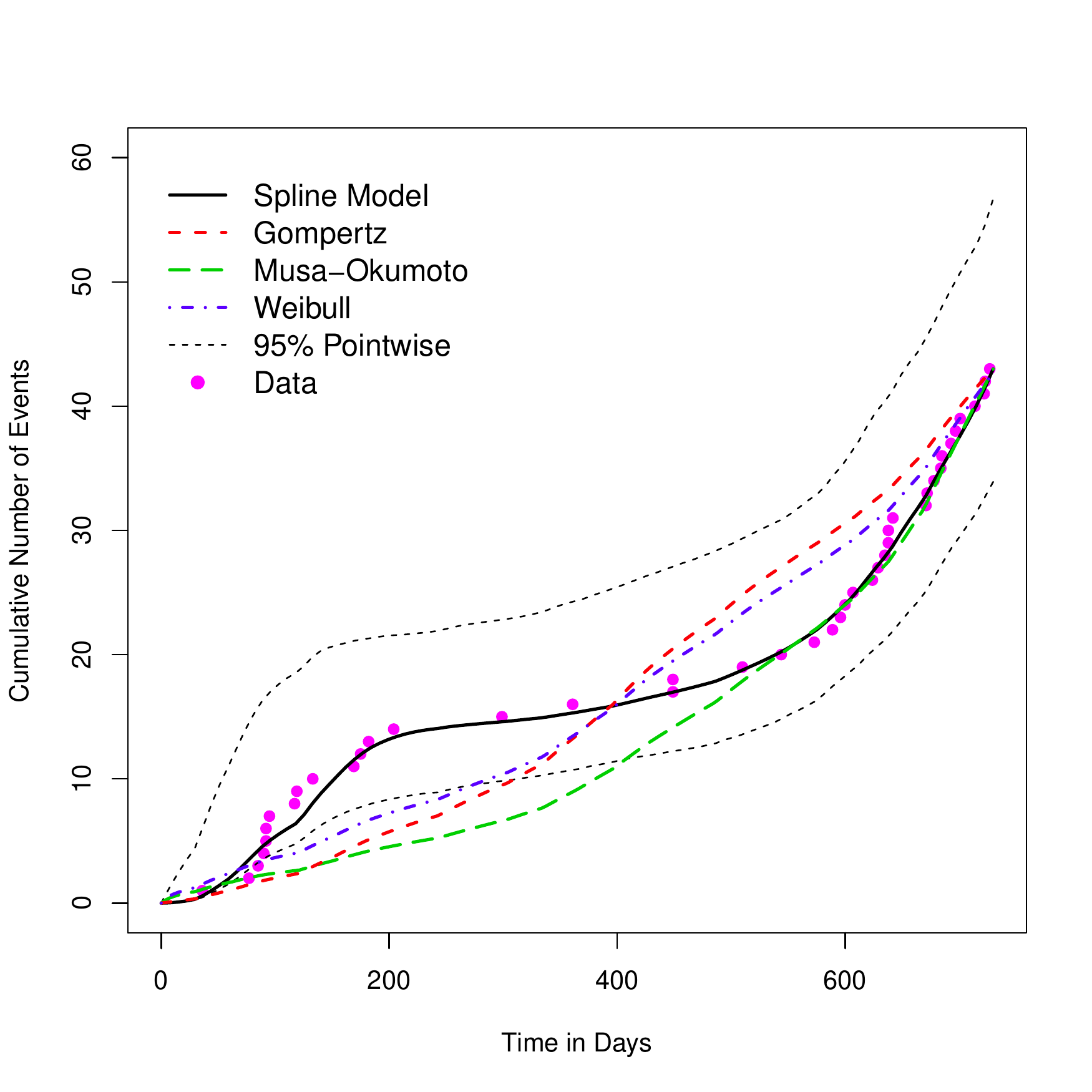}&
\includegraphics[width=.48\textwidth]{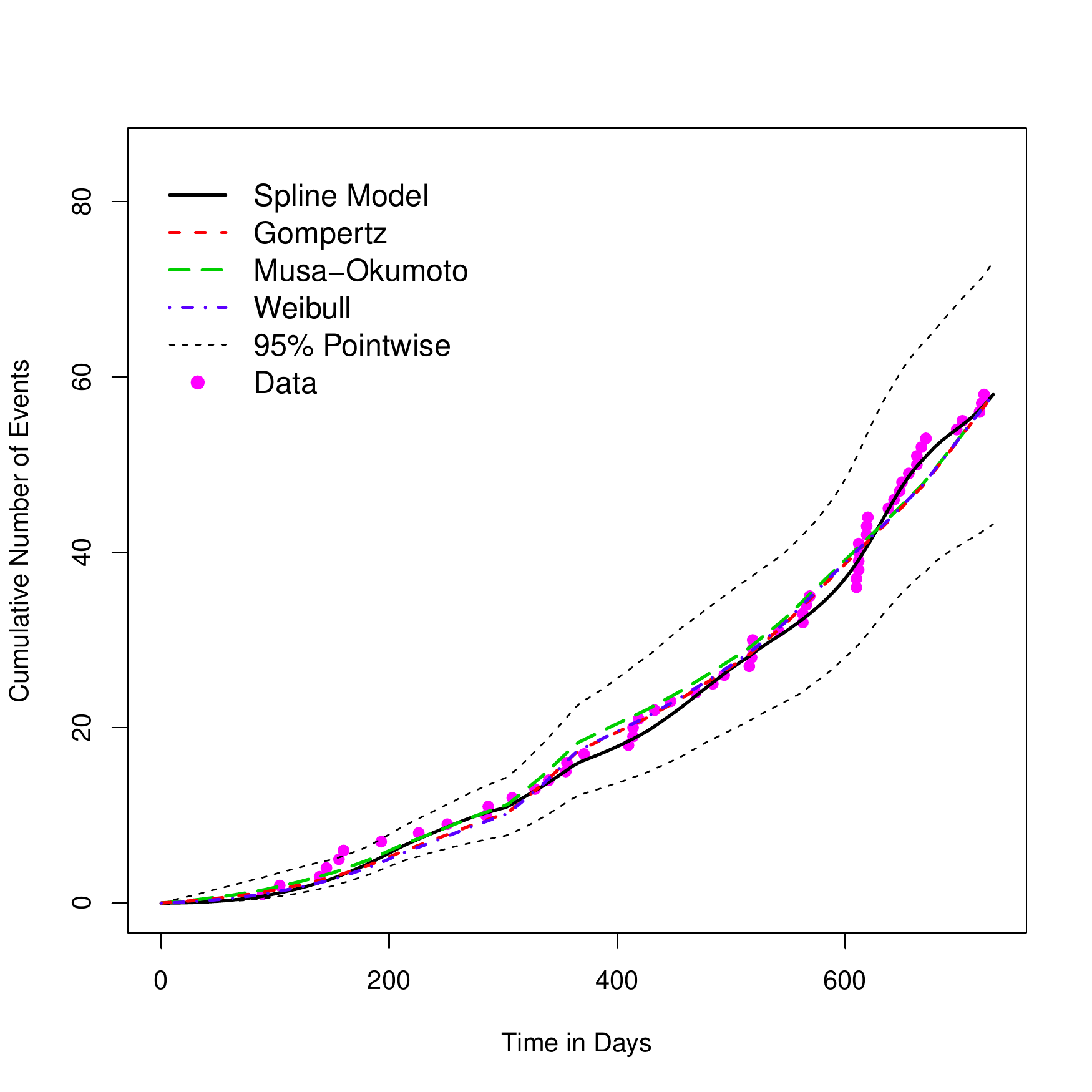}\\
(c) Pony AI & (d) Zoox\\
\end{tabular}
\caption{Plots of the expected versus the observed number of events for the four manufacturers based on the spline model and parametric models, together with the 95\% PCIs based on the spline model.}\label{fig:exp.vs.obs}
\end{figure}

To further check how well the models fit the data, Figure~\ref{fig:exp.vs.obs} shows the plots of the expected versus the observed number of events for the four manufacturers based on the spline model and parametric models, together with the 95\% PCIs based on the spline model. The expected number of events is computed based on the specific model with the adjustment for the mileage history from all units. The PCIs for the expected number of events are based on bootstrap samples. The shape of the function of the cumulative number of events differs from the shape of the BCIF because the function of the cumulative number of events is adjusted by the rate $x_i(t)$ which is time-varying and depends on the driving pattern, while BCIF is the case when $x_i(t)=1$. In all cases, the spline model tracks the cumulative number of observed events well. For Waymo, Cruise, and Zoox, the Gompertz and Weibull models also track the counts well, but visually we can see some departures for the Musa-Okumoto model. For Pony AI, all three parametric models show significant departures in the plot, indicating they are not flexible enough to describe that event process.

\subsection{Results Interpretation}
Table~\ref{tab:par.est} shows the parameter estimates, standard errors, and approximate 95\% confidence intervals (CIs) for the best parametric models for the four manufacturers. The CIs for parameters are based on a normal (Wald) approximation, and some transformations (e.g., a  logarithm transformation for positive parameters) are used to improve the performance. Note that the estimate for $\theta_1$ for the Zoox/Musa-Okumoto is set to zero because the model is degenerate at $\theta_1=0$, indicating a constant rate situation.

We also tested population heterogeneity using the procedure in Section~\ref{sec:frailty}. Table~\ref{tab:rand.eff} shows the summary of the gamma frailty models for the four manufacturers. All of the $p$-values are close to 1, indicating little population heterogeneity among the event processes for the four manufacturers. Hence, it is reasonable to use a model for which all units have the same event process with the same BCIF. Table~\ref{tab:rand.eff} also lists the names for the best parametric models for each manufacturer.

\begin{figure}
\centering
\includegraphics[width=.55\textwidth]{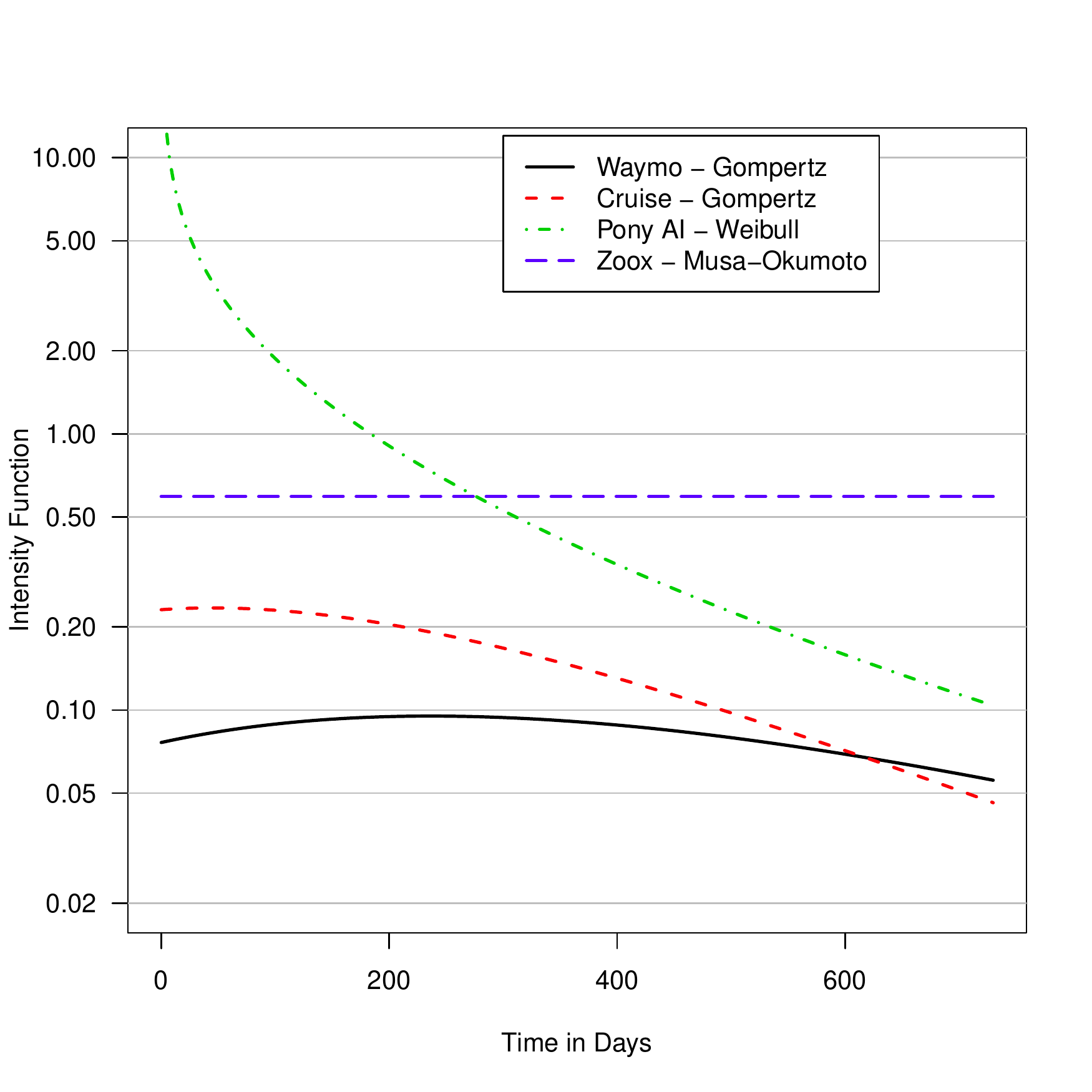}
\caption{Plot of BIFs from the best parametric models for the four manufacturers.}\label{fig:int.fun}
\end{figure}

To further visualize the results, Figure~\ref{fig:int.fun} plots the BIFs from the best parametric models for each of the four manufacturers. From the plot, we see the trends for different manufacturers. A decreasing trend means an improvement in AI technology and an increase in AV reliability. Waymo, Cruise, and Pony AI display a decreasing trend, while Zoox displays a constant rate of about 0.6 events per k-miles. Waymo starts at an event rate near 0.1 events per k-miles and decreases over the two-year period. Cruise starts at 0.2 events per k-miles and shows a decreasing trend. Pony AI starts at a high rate of 10 events per k-miles and shows a rapidly improving rate. By the end of the study period (i.e., November 30, 2019), the event rate for Waymo and Cruise is around 0.05 events per k-miles. The event rate for Pony AI is around 0.1 events per k-miles. This pattern indicates that there is a lot of improvement for the reliability of the Pony AI driving system. From the analysis conducted in this section for different manufacturers, we can see that the overall AV reliability is improving over the two-year period.

\begin{figure}
\centering
\begin{tabular}{cc}
\includegraphics[width=.48\textwidth]{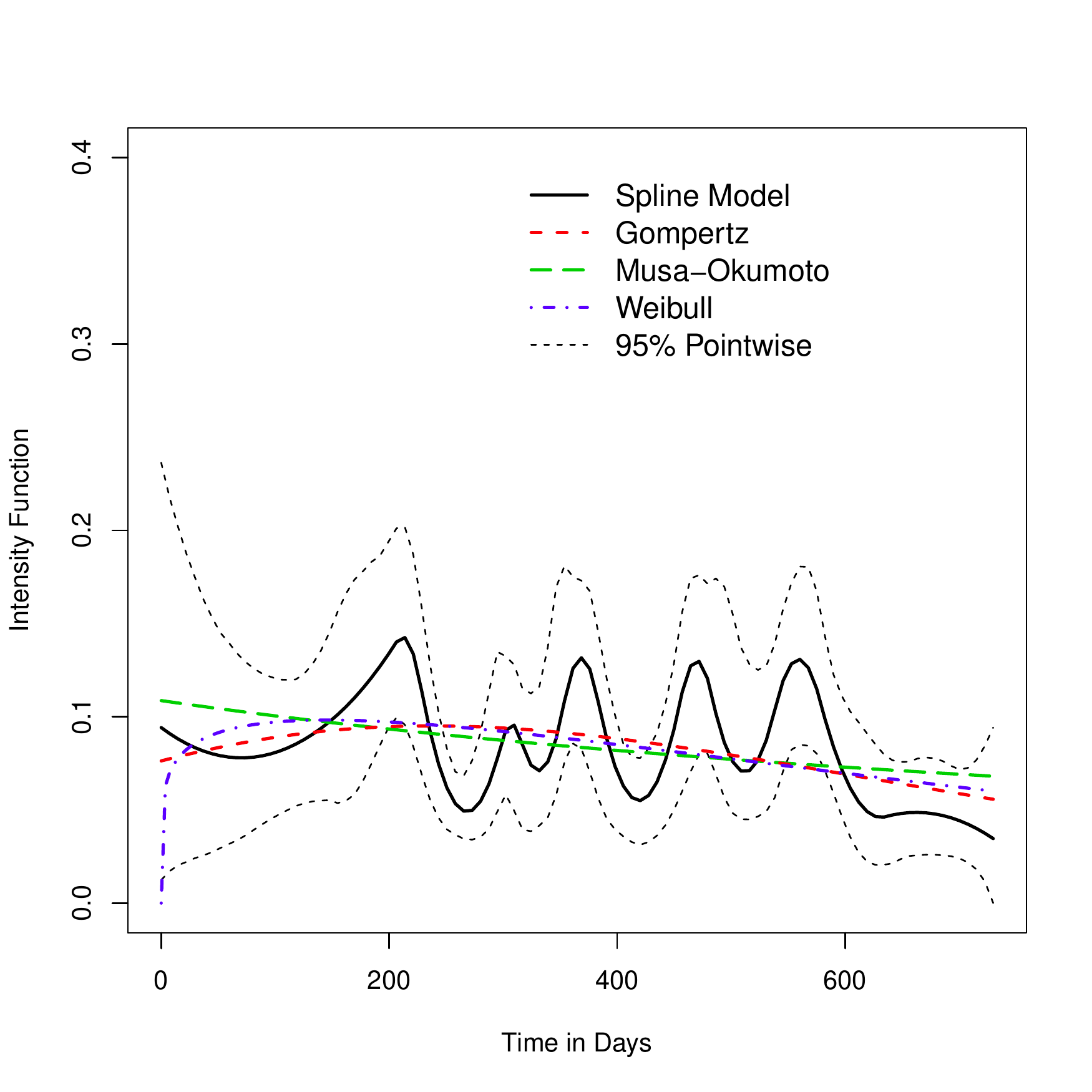}&
\includegraphics[width=.48\textwidth]{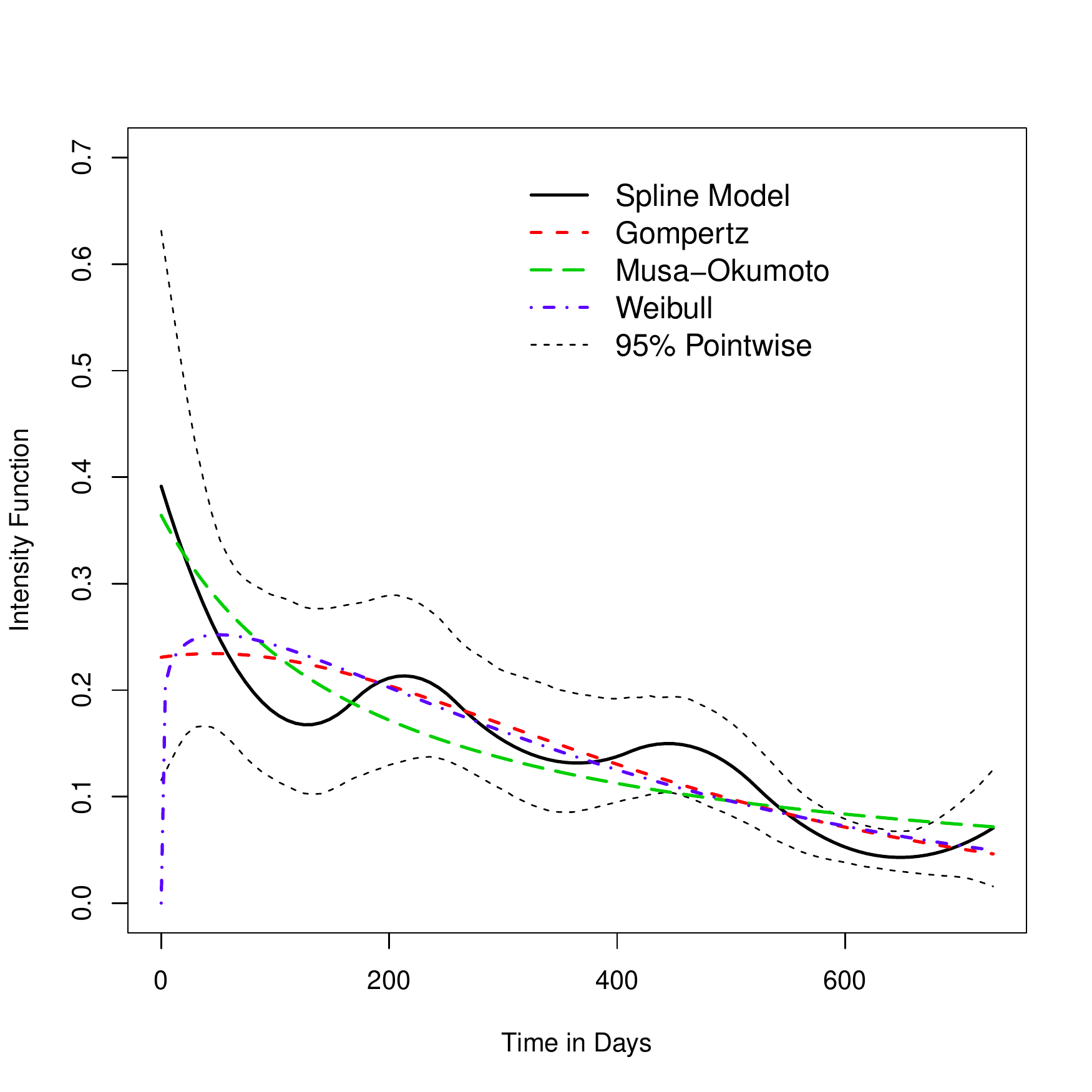}\\
(a) Waymo & (b) Cruise\\
\includegraphics[width=.48\textwidth]{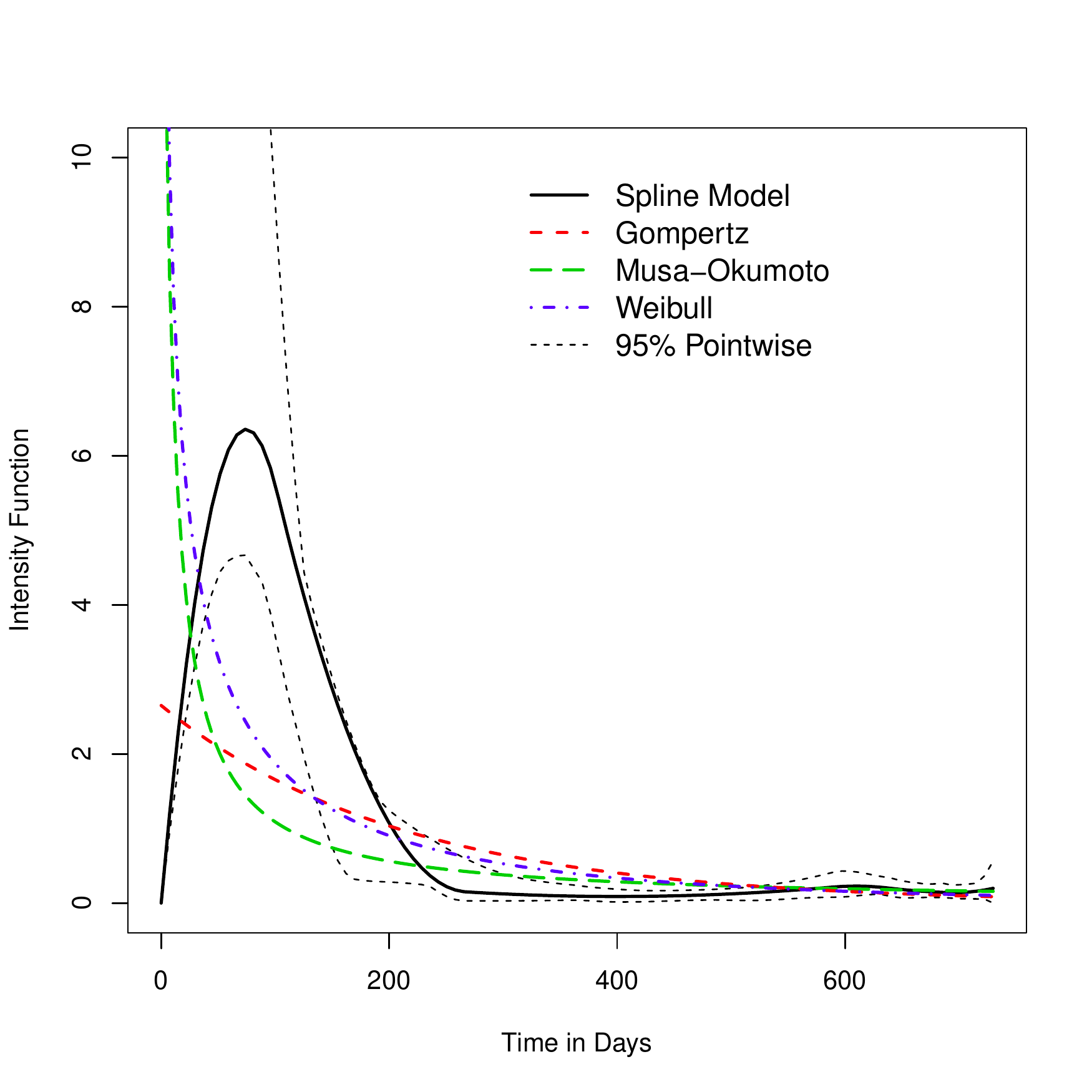}&
\includegraphics[width=.48\textwidth]{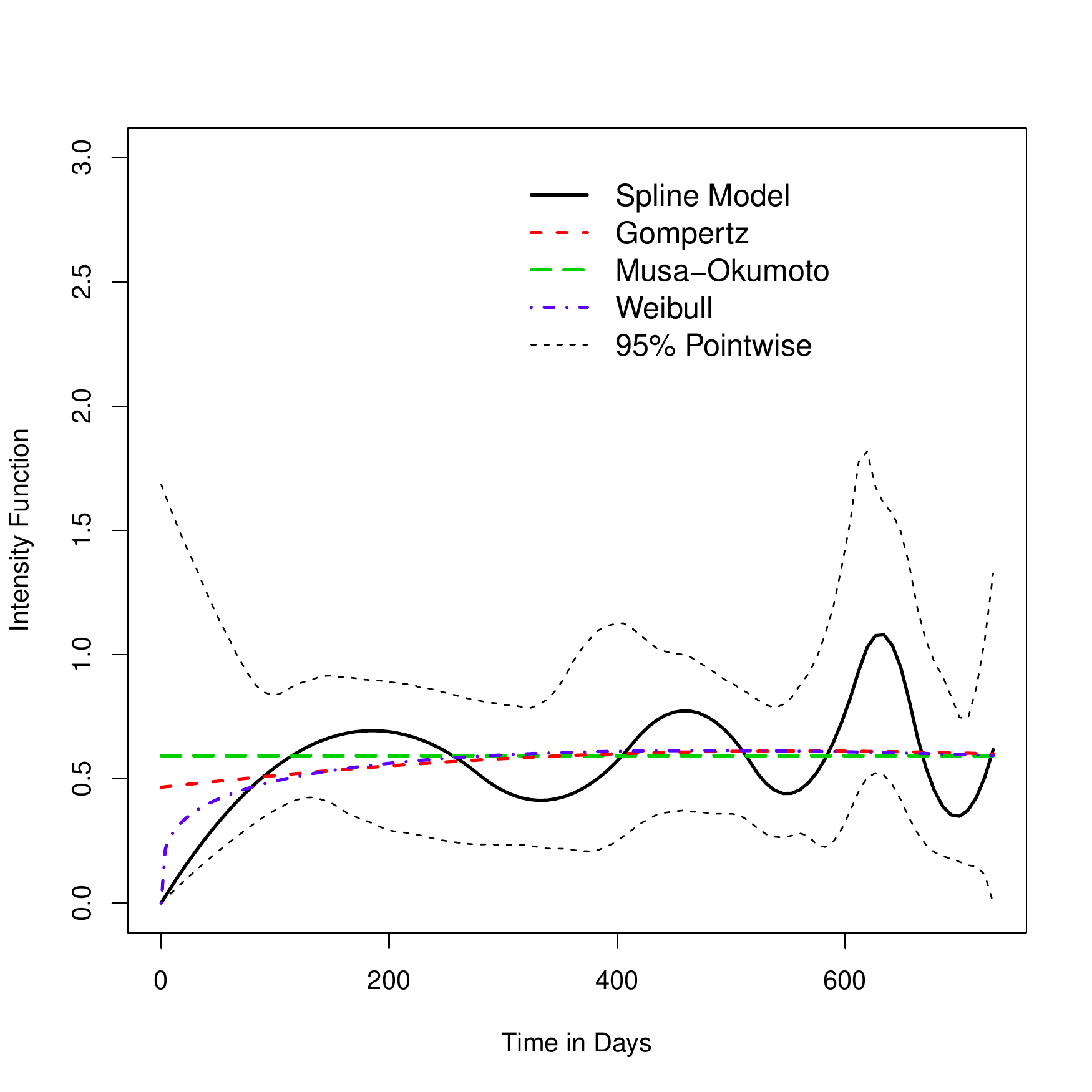}\\
(c) Pony AI & (d) Zoox\\
\end{tabular}
\caption{Plots of the estimated BIFs for the four manufacturers based on the spline model and parametric models, together with the 95\% PCIs based on the spline model.}\label{fig:int.est}
\end{figure}

As a comparison, Figure~\ref{fig:int.est} shows the estimated BIFs based on the spline model and the parametric models, and the 95\% PCIs based on the spline model. We can see that the spline model shows more variation but the general trends are the same as the parametric models. The estimates for Zoox have a large amount of variability due to the limited sample size, as discussed previously.

\begin{table}
\caption{Parameter estimates, standard errors, and approximate 95\% CIs for the best parametric models for the four manufacturers.}\label{tab:par.est}
\begin{center}
\begin{tabular}{c|c|rrrr}\hline\hline
Manufacturer/ &\multirow{2}{*}{Parameter} & \multirow{2}{*}{Estimate} & \multirow{2}{*}{Std. Err.} & \multicolumn{2}{c}{95\% CI}\\\cline{5-6}
Model       &     &  &  & Lower & Upper \\\hline
Waymo     & $\theta_1$ & 102.2539 &  31.7600 &  55.6278 &  187.9610 \\
 /        & $\theta_2$  &   0.9975 &   0.0009 &   0.9951 &    0.9987 \\
Gompertz  & $\theta_3$ &   0.1623 &   0.1229 &   0.0319 &    0.5326 \\\hline
Cruise    & $\theta_1$ & 171.3352 &  66.1936 &  80.3503 &  365.3472 \\
/         & $\theta_2$  &   0.9963 &   0.0009 &   0.9941 &    0.9977 \\
Gompertz  & $\theta_3$ &   0.3064 &   0.2285 &   0.0510 &    0.7842 \\\hline
Pony AI   & $\theta_1$ & 817.203 & 273.828 & 423.744 & 1575.997 \\
/         & $\theta_2$  &   0.0474 &   0.0474 &   0.0067 &    0.3363 \\
Weibull   & $\theta_3$ &   0.6304 &   0.1615 &   0.3815 &    1.0416 \\\hline
Zoox/     & $\theta_1$ &   0.0000 & 0.0000 & 0.0000 & 0.0000 \\
Musa-Okumoto & $\theta_2$ &   0.5933 &   0.0779 &   0.4587 &    0.7674 \\\hline\hline
\end{tabular}
\end{center}
\end{table}

\begin{table}
\caption{Summary of the gamma frailty models for the four manufacturers.}\label{tab:rand.eff}
\begin{center}
\begin{tabular}{c|ccc}\hline\hline\\[-2.5ex]
Manufacturer &  Best Parametric Model      & Variance ($\wh\phi$) & $p$-value \\ \hline
Waymo           & Gompertz      & 0.0001 & 0.9721   \\\hline
Cruise          & Gompertz      & 0.0000 & 0.9972   \\\hline
Pony AI         & Weibull       & 0.0000 & 0.9916   \\\hline
Zoox            & Musa-Okumoto  & 0.0197 & 0.8400   \\\hline\hline
\end{tabular}
\end{center}
\end{table}

\section{Concluding Remarks}\label{sec:conclusoin}

This paper focuses on the reliability analysis of AV technology using the recurrent disengagement events from the California driving study. We propose a statistical framework for modeling and analyzing the recurrent events data from AV driving tests. We use both parametric models and a nonparametric model to describe the event processes. Based on the spline model, we can select the best model, quantify uncertainty, and test heterogeneity in the event process. We want to point out that the parametric models and spline models are proposed as complementary tools for modeling and inference.

Through the simulation and data analysis, we show that the proposed spline model is flexible for describing the recurrent events data from four AV manufacturers, and parametric models are adequate for data from most manufacturers. It is worth noting that the best parametric models can be different for different manufacturers. The population heterogeneity in the event process is also low. From the data analysis, we found that the overall AV reliability is improving over the two-year study period.

The currently available data do not include covariates, such as the driving speed when the event occurred, the test environment (e.g., busy street versus freeway), and vehicle models. In the future, it would be interesting and useful to collect more covariate information. Our proposed modeling framework can be extended to analyze recurrent disengagement events data with covariates. The CA DMV recently started a driverless program where cars on the road do not have to have a driver. It would be interesting to analyze the driverless study data in the future when enough event data are available.

Because reliability is a property that evolves over time, all kinds of AI systems need to be tested over time to quantify their reliability. Although our analysis focuses on AV reliability, our proposed data analytic framework can also be applied to assess the reliability of other AI systems where departures from desired behavior provide recurrent events data. With an appropriate definition of time scale and events, the parametric and spline models discussed in this paper can be extended to analyze data from the reliability testing of other AI systems.

Computing hardware reliability is also an important aspect of AI reliability. For example, GPU's are widely used in AI model computing. \shortciteN{Ostrouchovetal2020} considered the lifetimes of GPUs used supercomputers. It would be interesting to study GPU reliability, or more broadly computing hardware reliability, and its relationship to AI reliability.

\section*{Acknowledgments}
The authors acknowledge the Advanced Research Computing program at Virginia Tech for providing computational resources. The work by Hong was partially supported by National
Science Foundation Grant CMMI-1904165 to Virginia Tech.



\end{document}